\documentclass[10pt, a4paper]{article}

\newcommand{\version}{final}
\usepackage[\version]{lrec2026} 

\usepackage{ifthen}
\newcommand{\footL}[2][]{\footnote{\scriptsize\urlstyle{tt}\url{#2}\ifthenelse {\equal {#1} {}}{}{\label{note:#1}}}}

\usepackage{multirow}
\usepackage{lipsum}

\usepackage{hyperref}

\newcommand{\tabh}[2]{\multicolumn{1}{#1}{\textbf{#2}}}
\newcolumntype{Z}{>{\raggedleft\arraybackslash}X}

\newif\ifshowappendix
\showappendixfalse

\makeatletter
\newcommand{\customlabel}[2]{%
  \protected@write\@auxout{}{\string\newlabel{#1}{{#2}{\thepage}{#2}{#1}{}} }%
}
\makeatother

\ifthenelse {\equal {\version} {review}}{  
\newcommand{\DataSetName}{\textsc{SiDiaC-v.2.0}}
\newcommand{\DataSetVer}{Version 2.0}
\newcommand{\OldDataSetName}{\textsc{SiDiaC}}

}{ 
\newcommand{\DataSetName}{\textsc{SiDiaC-v.2.0}}
\newcommand{\DataSetVer}{Version 2.0}
\newcommand{\OldDataSetName}{\textsc{SiDiaC-v.1.0}}

}

\usepackage{graphicx}

\usepackage{hhline}
\usepackage{makecell}
\usepackage{tabularx}
\usepackage{booktabs}
\usepackage{pifont}
\newcommand{\num}[1]{#1}

\newcommand{\LitCount}{80}
\newcommand{\DSTable}{ENGALL~\citep{davies2012expanding} & English & 5 & \phantom{-}1800~--~\phantom{-}1999 & $8.50\times10^{11}$ \\
ENGFIC~\citep{davies2012expanding} & English & 5 & \phantom{-}1800~--~\phantom{-}1999 & $7.50\times10^{10}$ \\
People in the News~\citep{hennig-wilson-2020-diachronic} & English & 5 & \phantom{-}2000~--~\phantom{-}2019 & $1.65\times10^{9\phantom{0}}$ \\
COHA~\citep{davies2012expanding} & English & 5 & \phantom{-}1810~--~\phantom{-}2009 & $4.10\times10^{8\phantom{0}}$ \\
EDGeS-English~\citep{bouma-etal-2020-edges} & English & 5 & \phantom{-}1301~--~\phantom{-}2020 & $3.28\times10^{8\phantom{0}}$ \\
English Scientific Writing~\citep{steuer-etal-2024-modeling} & English & 5 & \phantom{-}1665~--~\phantom{-}1996 & $2.96\times10^{8\phantom{0}}$ \\
Royal Society Corpus (RSC)~\citep{kermes-etal-2016-royal} & English & 5 & \phantom{-}1665~--~\phantom{-}1869 & $3.50\times10^{7\phantom{0}}$ \\
ARCHER~\citep{biber1994archer} & English & 5 & \phantom{-}1700~--~\phantom{-}1994 & $3.30\times10^{6\phantom{0}}$ \\
PPCHE-Modern British English~\citep{kroch2016penn} & English & 5 & \phantom{-}1707~--~\phantom{-}1914 & $2.80\times10^{6\phantom{0}}$ \\
PPCHE- Early Modern English~\citep{kroch2004penn} & English & 5 & \phantom{-}1500~--~\phantom{-}1720 & $1.70\times10^{6\phantom{0}}$ \\
PPCHE-Middle English~\citep{kroch2000penn} & English & 5 & \phantom{-}1150~--~\phantom{-}1500 & $1.20\times10^{6\phantom{0}}$ \\
ISWOC-English~\citep{bech2014iswoc} & English & 5 & \phantom{-}\phantom{0}400~--~\phantom{-}1010 & $2.94\times10^{4\phantom{0}}$ \\
GERALL~\citep{schneider1998adding} & German & 5 & \phantom{-}1800~--~\phantom{-}1999 & $4.30\times10^{10}$ \\
German Court Decisions~\citep{braun-2022-tracking} & German & 5 & \phantom{-}1970~--~\phantom{-}2020 & $2.70\times10^{10}$ \\
AMC~\citep{jutta2013linguistic} & German & 5 & \phantom{-}1986~--~\phantom{-}2012 & $1.05\times10^{10}$ \\
DTA~\citep{geyken2011deutsche} & German & 5 & \phantom{-}1600~--~\phantom{-}1899 & $1.55\times10^{8\phantom{0}}$ \\
EDGeS-German~\citep{bouma-etal-2020-edges} & German & 5 & \phantom{-}1301~--~\phantom{-}2020 & $9.03\times10^{7\phantom{0}}$ \\
ParlAT~\citep{wissik2018parlat} & German & 5 & \phantom{-}1945~--~\phantom{-}2017 & $7.50\times10^{7\phantom{0}}$ \\
RIDGES~\citep{odebrecht2017ridges} & German & 5 & \phantom{-}1478~--~\phantom{-}1870 & $3.00\times10^{6\phantom{0}}$ \\
ReM~\citep{klein2016handbuch} & German & 5 & \phantom{-}1050~--~\phantom{-}1350 & $2.00\times10^{6\phantom{0}}$ \\
GerManC~\citep{durrell2012germanc} & German & 5 & \phantom{-}1650~--~\phantom{-}1800 & $8.00\times10^{5\phantom{0}}$ \\
FREALL~\citep{sagot-etal-2006-lefff} & French & 5 & \phantom{-}1800~--~\phantom{-}1999 & $1.90\times10^{11}$ \\
LLODIA-French~\citep{armaselu-etal-2024-llodia} & French & 5 & \phantom{-}1690~--~\phantom{-}1918 & $6.40\times10^{6\phantom{0}}$ \\
SRCMF~\citep{lavrentiev2011syntactic} & French & 5 & \phantom{-}\phantom{0}800~--~\phantom{-}1299 & $2.51\times10^{5\phantom{0}}$ \\
ISWOC-French~\citep{bech2014iswoc} & French & 5 & \phantom{-}1225~--~\phantom{-}1275 & $2.34\times10^{3\phantom{0}}$ \\
CHIALL~\citep{xue2005penn} & Chinese & 5 & \phantom{-}1950~--~\phantom{-}1999 & $6.00\times10^{10}$ \\
ZhShiftEval~\citep{chen-etal-2022-lexicon} & Chinese & 5 & \phantom{-}1953~--~\phantom{-}2003 & $5.93\times10^{8\phantom{0}}$ \\
People’s Daily~\citep{he-etal-2014-construction} & Chinese & 5 & \phantom{-}1947~--~\phantom{-}1996 & $9.60\times10^{7\phantom{0}}$ \\
CORDE~\citep{shuger2020corpus} & Spanish & 5 & \phantom{-}1472~--~\phantom{-}1975 & $2.50\times10^{8\phantom{0}}$ \\
Corpus del Español (CdE)~\citep{davies2002corpus} & Spanish & 5 & \phantom{-}1200~--~\phantom{-}1999 & $1.00\times10^{8\phantom{0}}$ \\
CorDECh~\citep{contreras2009hacia} & Spanish & 5 & \phantom{-}1500~--~\phantom{-}1699 & $3.60\times10^{7\phantom{0}}$ \\
EZLN~\citep{gribomont-2023-diachronic-contextual} & Spanish & 5 & \phantom{-}1952~--~\phantom{-}2023 & $2.60\times10^{7\phantom{0}}$ \\
IAC-Spanish~\citep{sanchez-marco-etal-2010-annotation} & Spanish & 5 & \phantom{-}1100~--~\phantom{-}1599 & $2.00\times10^{7\phantom{0}}$ \\
IMPACT-es~\citep{sanchez2013open} & Spanish & 5 & \phantom{-}1482~--~\phantom{-}1990 & $8.00\times10^{6\phantom{0}}$ \\
PS Post Scriptum-Spanish~\citep{vaamonde2015ps} & Spanish & 5 & \phantom{-}1500~--~\phantom{-}1800 & $8.08\times10^{5\phantom{0}}$ \\
ISWOC-Spanish~\citep{bech2014iswoc} & Spanish & 5 & \phantom{-}1221~--~\phantom{-}1492 & $5.47\times10^{4\phantom{0}}$ \\
EDGeS-Dutch~\citep{bouma-etal-2020-edges} & Dutch & 4 & \phantom{-}1301~--~\phantom{-}2020 & $2.50\times10^{7\phantom{0}}$ \\
DiaCORIS~\citep{onelli-etal-2006-diacoris} & Italian & 4 & \phantom{-}1750~--~\phantom{-}1945 & $1.00\times10^{8\phantom{0}}$ \\
GDLI ~\citep{favaro-etal-2022-towards} & Italian & 4 & \phantom{-}1300~--~\phantom{-}1999 & $3.47\times10^{4\phantom{0}}$ \\
Kubhist~\citep{lilljegren2018introduktion} & Swedish & 4 & \phantom{-}1700~--~\phantom{-}1999 & $2.82\times10^{9\phantom{0}}$ \\
The Swedish Culturomics Gigaword corpus~\citep{eide2016swedish} & Swedish & 4 & \phantom{-}1950~--~\phantom{-}2015 & $1.00\times10^{9\phantom{0}}$ \\
EDGeS-Swedish~\citep{bouma-etal-2020-edges} & Swedish & 4 & \phantom{-}1301~--~\phantom{-}2020 & $8.30\times10^{6\phantom{0}}$ \\
FSV~\citep{delsing2002fornsvenska} & Swedish & 4 & \phantom{-}1276~--~\phantom{-}1734 & $1.20\times10^{6\phantom{0}}$ \\
Menota-Swedish~\citep{haugen2008menota} & Swedish & 4 & \phantom{-}1400~--~\phantom{-}1550 & $3.19\times10^{5\phantom{0}}$ \\
HaCOSSA~\citep{hoder2012annotating} & Swedish & 4 & \phantom{-}1375~--~\phantom{-}1550 & $1.28\times10^{5\phantom{0}}$ \\
MAþiR~\citep{text2024MApiR} & Swedish & 4 & \phantom{-}1200~--~\phantom{-}1299 & $3.30\times10^{4\phantom{0}}$ \\
CdP~\citep{davies2009creating} & Portuguese & 4 & \phantom{-}1200~--~\phantom{-}1900 & $4.50\times10^{7\phantom{0}}$ \\
Colonia~\citep{zampieri2013colonia} & Portuguese & 4 & \phantom{-}1500~--~\phantom{-}1999 & $6.20\times10^{6\phantom{0}}$ \\
TBCHP~\citep{galves2005change} & Portuguese & 4 & \phantom{-}1500~--~\phantom{-}1899 & $3.30\times10^{6\phantom{0}}$ \\
PS Post Scriptum-Portuguese~\citep{vaamonde2015ps} & Portuguese & 4 & \phantom{-}1500~--~\phantom{-}1800 & $6.48\times10^{5\phantom{0}}$ \\
ISWOC-Portuguese~\citep{bech2014iswoc} & Portuguese & 4 & \phantom{-}1344~--~\phantom{-}1400 & $3.64\times10^{4\phantom{0}}$ \\
DIAKORP~\citep{kuvcera2015diakorp} & Czech & 4 & \phantom{-}1300~--~\phantom{-}1999 & $4.00\times10^{6\phantom{0}}$ \\
DIALEKT~\citep{koprivova-etal-2014-mapping} & Czech & 4 & \phantom{-}1960~--~\phantom{-}1989 & $2.00\times10^{5\phantom{0}}$ \\
Menota-Norwegian~\citep{haugen2008menota} & Norwegian & 4 & \phantom{-}1200~--~\phantom{-}1350 & $8.67\times10^{5\phantom{0}}$ \\
RuSemShift~\citep{rodina-kutuzov-2020-rusemshift} & Russian & 4 & \phantom{-}1682~--~\phantom{-}2017 & $3.20\times10^{8\phantom{0}}$ \\
RuShiftEval~\citep{kutuzov-pivovarova-2021-three} & Russian & 4 & \phantom{-}1700~--~\phantom{-}2016 & $4.50\times10^{5\phantom{0}}$ \\
TOROT-Russian~\citep{eckhoff2015linguistics} & Russian & 4 & \phantom{-}1400~--~\phantom{-}1699 & $6.00\times10^{4\phantom{0}}$ \\
HGDS~\citep{simon2014corpus} & Hungarian & 4 & \phantom{-}\phantom{0}900~--~\phantom{-}1499 & $3.20\times10^{6\phantom{0}}$ \\
RoDICA~\citep{gifu2016tracing} & Romanian & 4 & \phantom{-}1840~--~\phantom{-}1991 & $4.71\times10^{5\phantom{0}}$ \\
Menota-Danish~\citep{haugen2008menota} & Danish & 3 & \phantom{-}1300~--~\phantom{-}1300 & $1.76\times10^{4\phantom{0}}$ \\
LLODIA-Hebrew~\citep{armaselu-etal-2024-llodia} & Hebrew & 3 & \phantom{-}1000~--~\phantom{-}2024 & $1.00\times10^{8\phantom{0}}$ \\
PROIEL-Ancient Greek~\citep{haug2008creating} & Greek & 3 & \phantom{-}\phantom{0}\phantom{0}\phantom{0}0~--~\phantom{-}\phantom{0}999 & $2.50\times10^{5\phantom{0}}$ \\
SLIEKKAS~\citep{gelumbeckaite2012senosios} & Lithuanian & 3 & \phantom{-}1500~--~\phantom{-}1800 & $3.50\times10^{5\phantom{0}}$ \\
IMP-sl~\citep{erjavec2015imp} & Slovene & 3 & \phantom{-}1584~--~\phantom{-}1918 & $3.00\times10^{5\phantom{0}}$ \\
PROIEL-Armenian~\citep{haug2008creating} & Armenian & 3 & \phantom{-}\phantom{0}400~--~\phantom{-}\phantom{0}450 & $2.35\times10^{4\phantom{0}}$ \\
LatinISE~\citep{mcgillivray2013tools} & Latin & 3 & \phantom{0}-186~--~\phantom{-}2000 & $1.30\times10^{7\phantom{0}}$ \\
PROIEL-Latin~\citep{haug2008creating} & Latin & 3 & \phantom{-}\phantom{0}300~--~\phantom{-}\phantom{0}499 & $2.25\times10^{5\phantom{0}}$ \\
GNC~\citep{gippert2015structuring} & Georgian & 3 & \phantom{-}\phantom{0}400~--~\phantom{-}2015 & $2.07\times10^{8\phantom{0}}$ \\
IcePaHC~\citep{rognvaldsson-etal-2012-icelandic} & Icelandic & 2 & \phantom{-}1100~--~\phantom{-}2012 & $1.00\times10^{6\phantom{0}}$ \\
Menota-Icelandic~\citep{haugen2008menota} & Icelandic & 2 & \phantom{-}1200~--~\phantom{-}1700 & $9.19\times10^{5\phantom{0}}$ \\
Greinir skáldskapar~\citep{eythorsson2014greinir} & Icelandic & 2 & \phantom{-}\phantom{0}900~--~\phantom{-}1270 & $3.14\times10^{4\phantom{0}}$ \\
Pre-Standard Irish~\citep{scannell-2022-diachronic} & Irish & 2 & \phantom{-}1600~--~\phantom{-}1936 & $3.80\times10^{3\phantom{0}}$ \\
SiDiaC~\citep{jayatilleke2025sidiac} & Sinhala & 2 & \phantom{-}\phantom{0}400~--~\phantom{-}1999 & $5.80\times10^{4\phantom{0}}$ \\
DCS~\citep{hellwig-etal-2020-treebank} & Sanskrit & 2 & -1300~--~\phantom{0}-700 & $5.40\times10^{5\phantom{0}}$ \\
FarPaHC~\citep{rognvaldsson-etal-2012-icelandic} & Faroese & 2 & \phantom{-}1800~--~\phantom{-}2012 & $5.30\times10^{4\phantom{0}}$ \\
DACON~\citep{o2024diachronic} & Newar & 1 & \phantom{-}1114~--~\phantom{-}1899 & $2.00\times10^{4\phantom{0}}$ \\
DIACU-Old Church Slavonic~\citep{cassese-etal-2025-diacu} & Slavonic & 1 & \phantom{-}\phantom{0}800~--~\phantom{-}1799 & $1.91\times10^{6\phantom{0}}$ \\
TOROT-Old Church Slavonic~\citep{eckhoff2015linguistics} & Slavonic & 1 & \phantom{-}\phantom{0}800~--~\phantom{-}1099 & $1.60\times10^{5\phantom{0}}$ \\
PROIEL-Old Church Slavonic~\citep{haug2008creating} & Slavonic & 1 & \phantom{-}\phantom{0}800~--~\phantom{-}1099 & $1.40\times10^{5\phantom{0}}$ \\
TOROT-Kiev-era Old East Slavic~\citep{eckhoff2015linguistics} & Slavonic & 1 & \phantom{-}\phantom{0}800~--~\phantom{-}1250 & $8.50\times10^{4\phantom{0}}$ \\
\hline}

\usepackage[nomessages]{fp} 

\FPeval{\TotalBookCount}{clip(233)}
\FPeval{\ScannedBookCount}{clip(221)}
\FPeval{\WrittenDateBookCount}{clip(59)}
\FPeval{\BookCount}{clip(185)}
\FPeval{\WordCount}{clip(229098)}

\FPeval{\FilteredWordCount}{clip(64805)}
\FPeval{\UniqWordCount}{clip(59331)}

\FPeval{\OldBookCount}{clip(46)}
\FPeval{\OldWordCount}{clip(58027)}
\FPeval{\OldFilteredWordCount}{clip(43959)}
\FPeval{\ConsistentKeys}{clip(80)}

\FPeval{\WordCountK}{clip(round(\WordCount/1000,0))}
\FPeval{\FilteredWordCountK}{clip(round(\FilteredWordCount/1000,0))}
\FPeval{\UniqWordPerCen}{clip(round((100*\UniqWordCount)/\WordCount,2))}

\usepackage[dvipsnames,table]{xcolor}
\usepackage{pifont}
\newcommand{\tick}{\textcolor{Black}{\ding{51}}}
\newcommand{\cross}{\textcolor{Black}{\ding{55}}}

\title{\DataSetName{}: Sinhala Diachronic Corpus \DataSetVer{}}

\name{Nevidu Jayatilleke$^\clubsuit$, Nisansa de Silva$^\clubsuit$, Uthpala Nimanthi$^\spadesuit$ \\ \large \textbf{Gagani Kulathilaka$^\spadesuit$,} \textbf{Azra Safrullah$^\spadesuit$,} \textbf{Johan Sofalas$^\spadesuit$}}

\address{$^\clubsuit$Department of Computer Science \& Engineering, University of Moratuwa, Sri Lanka \\
\texttt{\{nevidu.25, NisansaDdS\}@cse.mrt.ac.lk} \\
$^\spadesuit$Research Department, Informatics Institute of Technology, Sri Lanka \\
\texttt{\{uthpala.n, gagani.k, azra.s, johan.s\}@iit.ac.lk}}

\abstract{
\DataSetName{} is the largest comprehensive \textit{Sinhala Diachronic Corpus} to date, covering a period from 1800 CE to 1955 CE in terms of publication dates, and a historical span from the 5th to the 20th century CE in terms of written dates. The corpus consists of \num{\WordCountK}k words across \num{\BookCount{}} literary works that underwent thorough filtering, preprocessing, and copyright compliance checks, followed by extensive post-processing. Additionally, a subset of \WrittenDateBookCount{} documents totalling \num{\FilteredWordCountK}k words was annotated based on their written dates. Texts from the \textit{National Library of Sri Lanka} were selected from the \OldDataSetName{} non-filtered list, which was digitised using \texttt{Google Document AI OCR}. This was followed by post-processing to correct formatting issues, address code-mixing, include special tokens, and fix malformed tokens. The construction of \DataSetName{} was informed by practices from other corpora, such as \texttt{FarPaHC}, \OldDataSetName{}, and \texttt{CCOHA}. This was particularly relevant for syntactic annotation and text normalisation strategies, given the shared characteristics of low-resource language status between Faroese and the similar cleaning strategies utilised in \texttt{CCOHA}. This corpus is categorised into two layers based on genres: primary and secondary. The primary categorisation is binary, assigning each book to either Non-Fiction or Fiction. The secondary categorisation is more detailed, grouping texts under specific genres such as Religious, History, Poetry, Language, and Medical. Despite facing challenges due to limited resources, \DataSetName{} serves as a comprehensive resource for Sinhala NLP, building upon the work previously done in \OldDataSetName{}. 
\\ \newline 
\Keywords{Sinhala, Diachronic Corpus, Low-resource Languages}
}

\begin{document}

\maketitleabstract

\section{Introduction}

\raisebox{-0.5ex}{%
    \includegraphics[height=1.45\fontcharht\font`\A]{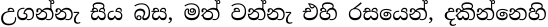}%
} \raisebox{-0.5ex}{%
    \includegraphics[height=1.45\fontcharht\font`\A]{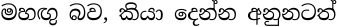}}\footnote{\scriptsize\textbf{English Meaning:} \textit{Learn your mother tongue, be intoxicated by its taste; the preciousness you see in it, enlighten others of} - \texttt{Kumaratunga Munidasa}.\\ \textbf{Sinhala to IPA Transliteration:} \raisebox{-0.5ex}{%
    \includegraphics[height=1.35\fontcharht\font`\A]{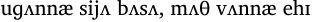}%
} \raisebox{-0.5ex}{%
    \includegraphics[height=1.35\fontcharht\font`\A]{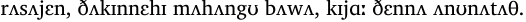}%
}}; highlights the cultural reverence for linguistic identity and the enduring significance of one’s native language across generations. Languages are constantly evolving, changing over time at all levels of linguistic structure. These changes are influenced by external factors, such as cultural shifts and technological advancements~\cite{alatrash2020ccoha, blank1999new, fromkin2017introduction}. The field of historical or diachronic linguistics focuses on the study and analysis of how languages change over time. In the past two decades, researchers have shown a growing interest in various aspects of diachronic language change. This increased attention can be attributed to technological advances, including the digitisation of historical texts, improvements in computational power, and the availability of large-scale historical corpora specifically designed for diachronic studies~\cite{alatrash2020ccoha, tahmasebi2018survey, tang2018state, bowern-2019-semantic}.

\begin{figure*}[!htbp]
    \centering
    \includegraphics[width=\textwidth]{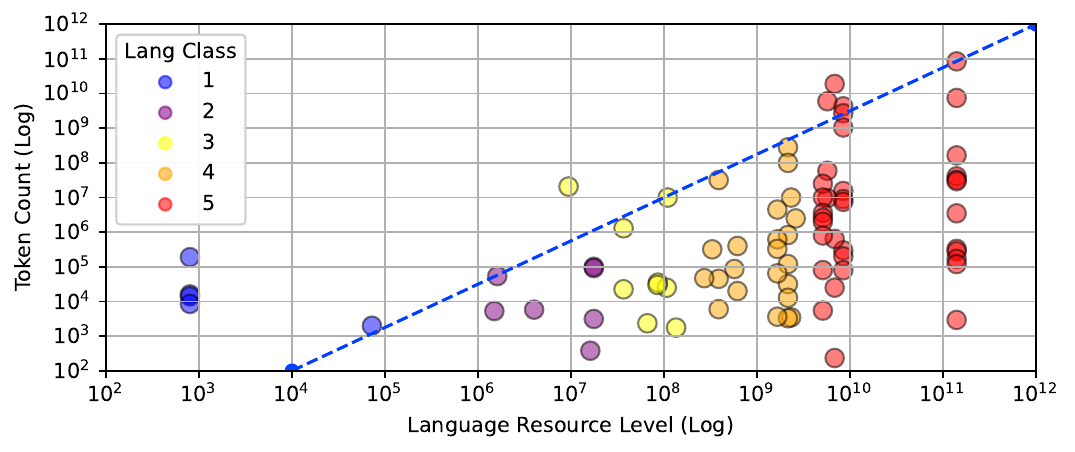}%
    \caption{Log token counts of the existing Diachronic corpora against the log of language resource level}
    \label{fig:Copora}
\end{figure*}

The Sinhala language is an Indo-European language with a rich and diverse literary heritage that has developed over several millennia. Its origins can be traced back to between the 3rd and 2nd centuries BCE. Unlike English, which is part of the Germanic branch, Sinhala belongs to the Indo-Aryan branch of the Indo-European language family. The Sinhala language, which is the primary language of the Sinhalese people who constitute the largest ethnic group in the island nation of Sri Lanka, is recognised as the first language (L1) for about 16 million individuals~\cite{de2025survey}. Furthermore, while English uses the Latin alphabet, Sinhala has its own writing system, which is a descendant of the Indian Brahmi script~\cite{fernando1949palaeographical, de-mel-etal-2025-sinhala}. Sinhala is classified as a lower-resourced language (Category \num{02}) according to the criteria presented by~\citet{ranathunga-de-silva-2022-languages}.


In this study, we introduce the largest \textit{Sinhala Diachronic Corpus} to date, \DataSetName{}\footL[sdc]{https://github.com/NeviduJ/SiDiaC-v.2.0}, which encompasses the period from 1800 CE to 1955 CE based on publication dates, and a range from the 5th century to the 20th century based on written dates. We provide a detailed discussion on the creation of the entire corpus, starting with data collection, preprocessing, and filtration, followed by digitisation and extensive post-processing, and concluding with metadata creation. Additionally, we present an evaluation of both the corpus and the metadata, as well as a century-wise Bag of Words (BoW) analysis conducted on a subset of \DataSetName{}.

\section{Existing Work}
\label{sec:Existing}



It is important to note that existing corpora vary significantly in size. For example, the \textit{Faroese Parsed Historical Corpus} (\texttt{FarPaHC})~\cite{rognvaldsson-etal-2012-icelandic} contains about 53,000 words, while the \textit{Corpus of Historical American English} (\texttt{COHA})\footL[coha]{https://www.english-corpora.org/coha/}~\cite{davies2012expanding} has approximately 400 million words. In contrast, the \textit{Google Books Ngram Corpus}\footL[gngram]{https://books.google.com/ngrams/}~\cite{lin-etal-2012-syntactic} comprises billions of words. The size of a corpus is influenced not only by the textual resources available for a given language, but also by the level and quality of annotation provided within the corpus~\cite{pettersson2022swedish}.

In Figure~\ref{fig:Copora}, we show the log token counts of \num{\LitCount{}} existing Diachronic corpora\footnote{\scriptsize Details available on Table~\ref{tab:AllDS} of Appendix~\ref{app:AllDS}.} against the log of language resource level calculated by~\citet{ranathunga-de-silva-2022-languages}. The language class is also taken from~\citet{ranathunga-de-silva-2022-languages}, which is based on the initial categorisation of~\citet{joshi-etal-2020-state}. We observe that 10\textsuperscript{-2} of the resource level proposed by~\citet{ranathunga-de-silva-2022-languages} seems to be the soft-cap for the size of the diachronic corpora for any language class, as shown by the dashed line. Note the extremely low-resourced language \textit{Slavonic} punching way above its weight due to it being included in \texttt{DIACU}~\cite{cassese-etal-2025-diacu}, \texttt{PROIEL}~\cite{haug2008creating}, and \texttt{TOROT}~\cite{eckhoff2015linguistics}. The \texttt{GNC} corpus~\cite{gippert2015structuring} pushes \textit{Georgian} above the line. This uneven resource availability for \textit{Georgian} also reflects from~\citet{ranathunga-de-silva-2022-languages} categorising it as class 3, although by total resource count it is ostensibly in the range of class 2. 
A comprehensive literature survey, as well as details on all \LitCount{} corpora are discussed in Appendix~\ref{app:AllDS}. 

\subsection{\OldDataSetName}
\label{subsec:sidiac_lit}

The \textit{Sinhala Diachronic Corpus} (\OldDataSetName) is the first comprehensive diachronic corpus for the Sinhala language, covering a historical period from the 5th to the 20th century CE, specifically from 426 CE to 1944 CE. It comprises \num{\OldWordCount} word tokens extracted from \num{\OldBookCount} literary works~\citelanguageresource{SiDiaC}.

The construction of \OldDataSetName{} involved several detailed steps. First, Sinhala literature was primarily sourced from the \textit{National Library of Sri Lanka}\footL[natlib]{https://www.natlib.lk/}. Next, a rigorous data filtration process was applied, which considered the availability of scanned copies, the accurate determination of written dates (distinct from issue dates), and compliance with Sri Lankan copyright laws\footL[IPAct]{https://www.gov.lk/wordpress/wp-content/uploads/2015/03/IntellectualPropertyActNo.36of2003Sectionsr.pdf}. This process reduced the initial selection of \num{\TotalBookCount} identified books down to the final \num{\OldBookCount}.

Text extraction was carried out using \texttt{Google Document AI}\footL[document-AI]{https://cloud.google.com/document-ai/} OCR, chosen for its superior real-world accuracy~\cite{jayatilleke-de-silva-2025-zero}. The OCR technology also demonstrated advanced capabilities, such as text modernisation 
and morpheme segmentation for historical Sinhala. 
Although the average OCR accuracy was \num{96.84}\%, extensive manual post-processing was conducted by native Sinhala-speaking authors using a human-in-the-loop strategy to correct various formatting issues. These included spacing errors, multi-column text, misplaced words and phrases, paragraph and line indentation issues, and the removal of seals and page numbers. \OldDataSetName{} serves as a foundational resource for Sinhala Natural Language Processing, enabling diachronic linguistic studies.


\subsection{\texttt{COHA} \& \texttt{CCOHA}}


\texttt{COHA}, released in late 2010, is a crucial linguistic resource comprising about \num{400} million words from over \num{100000} texts published between the 1810s and 2009~\cite{davies2012expanding}. It is genre-balanced across decades to accurately reflect linguistic changes. The construction of \texttt{COHA} involved assembling texts from various archives, scanning books with OCR, and converting over \num{40000} newspaper PDF files. 
The corpus was then lemmatised and Part-of-Speech (PoS) tagged using the \texttt{CLAWS}\footL[claws]{https://ucrel.lancs.ac.uk/claws/} tagger~\cite{rayson1998claws}. 

The downloadable versions of \texttt{COHA}~\citelanguageresource{COHA} had several limitations, including special `\texttt{@}' tokens that disrupted token flow and obscured sentence boundaries. Issues such as malformed tokens, "\texttt{NUL}" control characters, inconsistent lemmas, incorrect PoS tags, and escaped HTML characters complicated sentence-level Natural Language Processing tasks. To overcome these challenges while preserving core properties, the \textit{Clean Corpus of Historical American English} (\texttt{CCOHA})\footL[ccoha]{https://www.ims.uni-stuttgart.de/forschung/ressourcen/korpora/ccoha/} was developed through a two-pass cleaning process using \texttt{Python} and \texttt{NLTK}\footL[nltk]{https://www.nltk.org/}~\cite{alatrash2020ccoha}.

The first pass addressed initial errors by replacing "\texttt{NUL}" control characters, removing non-word tokens, unescaping HTML characters, unifying lemmas, and marking malformed tokens. The second pass focused on contextual cleaning, tagging and lemmatising tokens using full sentence context and mapping \textit{Penn Treebank} PoS tags to \texttt{CLAWS7}\footL[claws7]{https://ucrel.lancs.ac.uk/claws7tags.html} tags for clarity. These extensive cleaning efforts resulted in significant improvements: \texttt{CCOHA}~\citelanguageresource{CCOHA}, which contains over 25 million more word tokens and nearly two million more non-word tokens.

\section{Limitations of \OldDataSetName}
\label{subsec:limits_sidiac}

We have identified several limitations of \OldDataSetName{}, some of which were acknowledged by the authors of the corpus~\cite{jayatilleke2025sidiac}. Additionally, we discovered further issues during the analysis, as listed below:


\vspace*{-3.5mm}
\paragraph{Data Filtration:} Out of the \ScannedBookCount{} scanned copies acquired for \OldDataSetName, only 59 had written dates or periods identified. From these 59, 13 were removed to comply with copyright laws, leaving just \OldBookCount{} books. The written dates were annotated based on the lifespans of well-known authors, while the majority of the remaining dates heavily relied on the work of~\citet{Sannasgala_2009}, indicating an over-reliance on a single source. This shows the difficulty in identifying written dates and indicates that the current strict criteria for filtering data may need to be revised to include a broader range of literature.

\vspace*{-3.5mm}
\paragraph{Malformed Tokens:} During the post-processing of \OldDataSetName, only five formatting issues were addressed: errors in word and character spacing, multi-column text rendering issues, misplaced words and phrases, incorrect paragraph and line indentations, and the removal of seal context and page numbering. However, the most pertinent potential word or character-level identification errors identified from the OCR process were not addressed. These errors can be categorised as spell-correction domain errors, where substitutions, deletions, and insertions of characters and words occurred mainly due to noise in the scanned document, which led to misinterpretations by the OCR engine.


\vspace*{-3.5mm}
\paragraph{Code-Mixed Data:} It was noted that \OldDataSetName{} contains Pali, Sanskrit, and English mixed with Sinhala. Additionally, we identified certain books, such as \textit{Adhimasa Dheepanaya}, which are entirely in Pali. This raises concerns about whether the corpus is fully representative of Sinhala literature. 

\vspace*{-3.5mm}
\paragraph{Commentary Books:} In \OldDataSetName, there are books with titles that include the word "\raisebox{-0.35ex}{%
    \includegraphics[height=1.5\fontcharht\font`\A]{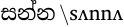}%
}". These books are commentaries on the previous works. As a result, the textual contents span two different time periods, even though the annotations refer to the original book's date of composition. As an example, for \textit{Sanna sahitha Salalihini Sandeshaya}, the declared written date in \OldDataSetName{} pertains to the original \textit{Salalihini Sandeshaya}; but the majority of the text is in fact commentaries which were written long after the original text.

\begin{figure*}[!htbp]
    \centering
        \includegraphics[width=0.95\textwidth]{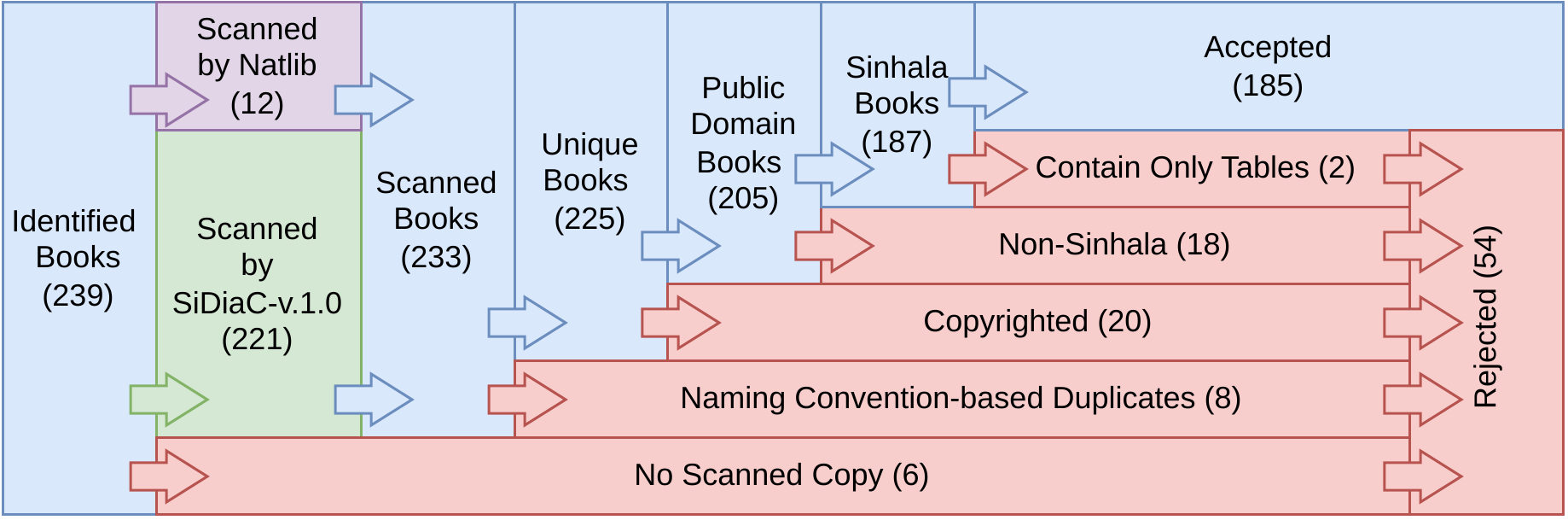}%
    \caption{Sequential Data Filtration Procedure}
    \label{fig:datafiltration}
\end{figure*}

\vspace*{-3.5mm}
\paragraph{Poetry Suffixes:} The poems in both \OldDataSetName{} and \DataSetName{} emphasise rhyming through the use of "\raisebox{-0.5ex}{%
    \includegraphics[height=1.45\fontcharht\font`\A]{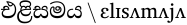}%
}" and "\raisebox{-0.5ex}{%
    \includegraphics[height=1.45\fontcharht\font`\A]{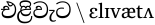}%
}," where the rhyming sound is isolated or detached from the full word. "\raisebox{-0.5ex}{%
    \includegraphics[height=1.45\fontcharht\font`\A]{Figures/si_046.pdf}%
}" ensures that the rhyme sound at the end of each line, while "\raisebox{-0.5ex}{%
    \includegraphics[height=1.45\fontcharht\font`\A]{Figures/si_047.pdf}%
}" highlights this separated sound at the start and\textbackslash or middle of each line through repetition and character isolation from the related word, enhancing the metrical structure and auditory appeal~\cite{kumara2017sri}.



\vspace*{-3.5mm}
\paragraph{Page Titles:} Due to the fact that \OldDataSetName{} was created by only including the initial five to eight pages of literary works, during the analysis of the textual context extracted using OCR, we observed that the first pages usually contain the book title and the chapter title with the main text of the book. 
Additionally, most pages were found to contain the book name or the chapter name in the header. This title information, which does not hold contextual relevance for the language studies that this corpus enables, was not removed in \OldDataSetName.

\vspace*{-3.5mm}
\paragraph{Multi-Column Text:} During the post-processing of the text extracted using OCR in \OldDataSetName{}, an issue was identified where the OCR engine failed to recognise the two columns separately, resulting in incorrectly rendered text. The proposed solution was to represent the data in a two-column format to replicate the layout of the original books. However, this approach has significant limitations when retrieving information, even from text files. Processing text in this format makes it challenging to effectively separate the two columns for downstream applications.

\vspace*{-3.5mm}
\paragraph{Content Tables:} During our review of the text files in comparison to the corresponding PDF files in \OldDataSetName, we observed that some content tables attempted to replicate their format by indenting text instead of using line separations. Additionally, letters or words in the table format were recreated in the text content by inserting multiple spaces between them. This structure is not suitable for retrieving information from tables, particularly when working with code scripts.

\vspace*{-3.5mm}
\paragraph{Footnotes:} During our review of the PDF files in \OldDataSetName, we observed that the inclusion of footer content, such as definitions of terminologies or related information, was common. However, adding these text fragments into a flat text file could disrupt the overall flow, as there are no page separations or footer indicators to contextualise this information, unlike the visual layout found in PDF files. 

\vspace{0.5em}
\noindent
The limitations identified in \OldDataSetName{} were discussed further, with supporting evidence provided in Appendix~\ref{app:limits_sidiac}.

\section{Methodology}

In this section, we will discuss the outline of the procedures implemented in \DataSetName{} to address the limitations of \OldDataSetName. Additionally, we will present the process of expanding the dataset from \OldBookCount{} literary works to \BookCount, utilising careful filtration mechanisms to ensure quality. The complete filtration pipeline is shown in Figure~\ref{fig:datafiltration}.


\subsection{Revised Written-Date Filtering}

In \DataSetName, significant emphasis was placed on expanding the collection to include more literary works than in \OldDataSetName. \OldDataSetName{} initially considered \TotalBookCount{} unique books, but after undergoing multiple filtration steps, only \OldBookCount{} were finalised to be included in the collection. A crucial step in this process involved applying a written date-based filter, which eliminated 168 out of the original \TotalBookCount{} books from the identified literature. 
However, after examining other diachronic corpora, we realised that the filtration criteria of \OldDataSetName{} were overly strict. While some corpora, such as \texttt{PROIEL}~\cite{haug2008creating}, \texttt{TOROT}~\cite{eckhoff2015linguistics}, \texttt{IcePaHC}~\cite{rognvaldsson-etal-2012-icelandic}, \texttt{DCS}~\cite{hellwig-etal-2020-treebank}, \texttt{DACON}~\cite{o2024diachronic}, \texttt{LatinlSE}~\cite{mcgillivray2013tools} do use manuscript (composition) dating, others, such as \texttt{COHA}~\cite{davies2012expanding}, Royal Society Corpus (\texttt{RSC})~\cite{kermes-etal-2016-royal}, \texttt{EDGeS}~\cite{bouma-etal-2020-edges}, \texttt{ParlAT}~\cite{wissik2018parlat}, generally uses the publication year,
%
while \texttt{COHA} also takes into account the authors' lifespans when applicable. As a result, we decided to adopt a similar approach to \texttt{COHA} for our annotations, enabling us to include a broader range of literary works in \DataSetName.

\subsection{Data Preparation for OCR}
\label{subsec:data_prep}

Initially, we obtained a complete list of \TotalBookCount{} books from \OldDataSetName{} (prior to their filtration), which they sourced from the digital library of \texttt{Natlib}. However, upon our initial inspection of these books, we discovered that 8 of them were duplicates listed separately only due to slight differences in the naming of the files. As a result, we were left with 225 unique books.

Afterwards, we renamed the books according to the metadata list, as many had inconsistent file names that made identification difficult (e.g., some had only the first word of the title, others included an identifier number, and some had titles in romanised Sinhala). We then inspected all PDF files to identify any issues that needed to be addressed and found that 12 books had all their pages available. These are the books that \texttt{Natlib} has made openly accessible to anyone through the digital library prior to the creation of \OldDataSetName. Including all pages of these books would provide us with a larger number of tokens, but this could lead to token bias toward those specific literary sources, as the other remaining books typically only have 5 to 8 pages digitised. Therefore, we decided to include the first 15 pages of these books to maximise the number of tokens while avoiding bias in our corpus.

Furthermore, we identified books that had been scanned with a 90-degree counterclockwise rotation, which had resulted in being excluded from \OldDataSetName. Additionally, we found that some books have been scanned in a single landscape A4 view, where two portrait pages are scanned into one landscape PDF page, which has resulted in these books to suffer from the multi-column rendering error in \OldDataSetName{} despite not inherently being multi-column books. We cropped and sorted these pages correctly to prevent such issues arising in \DataSetName. 


\subsection{Copyright and Language Filtration}
\label{subsec:copyright_lang_filter}


Copyright laws in Sri Lanka are governed by the \texttt{Intellectual Property Act No. 36 of 2003}\footref{note:IPAct}. According to this act, copyright protection generally lasts for the lifetime of the author, plus an additional 70 years after their death. In cases where the author is unknown, copyright protection extends for 70 years from the date of first publication. Therefore, we focused on literature by authors who passed away before 1955, as well as works by unknown authors published before that year. After applying this criterion, we had to remove 20 documents from our list.

Following that, we removed documents that contained entirely non-Sinhala text and those that contained only tabular information. The detailed procedures for these filtration steps are outlined in subsections~\ref{subsec:code_mix}, \ref{subsec:other_post_steps}, and Appendix~\ref{app:limits_sidiac}. As a result of this filtration process, we removed a total of 20 documents: 18 that were non-Sinhala and 2 that contained only tables, from the initial set of 205 documents that remained after the naming convention-based duplicates removal, finally resulting in \BookCount{} documents.

\subsubsection{Written Date Annotation}
\label{subsec:writte_date}

The issue date of the identified books was clearly indicated in the digital repository at \texttt{Natlib}. However, this does not necessarily mean that the books were actually written during those specified dates. In fact, a book could have been written centuries earlier, while its printed version was released much later. Document dating has become widely recognised in computational sociology and studies within the digital humanities~\cite{ren-etal-2023-time,baledent-etal-2020-dating,hellwig-2020-dating}. Compared to other dating tasks, dating historical texts is more complex due to the lack of explicit temporal indicators (such as time expressions) that would help determine when a document was written~\cite{toner2019language,baledent-etal-2020-dating,hellwig-2020-dating}.

It is evident that text dating or the procedure of annotating the written date of a document is a crucial task in diachronic studies~\cite{ansari2023diachronic, ren-etal-2023-time,favaro-etal-2022-towards}. While there are diachronic corpora that retain only the issue date~\cite{gribomont-2023-diachronic-contextual,he-etal-2014-construction}, we conducted a thorough analysis to ensure that the actual written year or possible time period of the books was accurately represented when applicable. A similar approach was adopted in \texttt{COHA}, where they included the author's lifespan when applicable~\cite{davies2012expanding}.


For the identification process, we used the book by~\citet{Sannasgala_2009}, similar to \OldDataSetName. We also found a Sinhala dictionary by \citet{Soratha_Godage} that listed the written periods in centuries of many books used to create the respective dictionary, helping us identify overlaps and anchor certain texts. Furthermore, when applicable, we annotated documents based on the lifespan of the author. However, we could only do this for well-known authors whose information is still present in historical records.

\subsection{Text Extraction using OCR}
\label{subsec:text_extract}

\begin{figure}[!htbp]
    \centering
    \setlength{\fboxsep}{1pt} 
    \setlength{\fboxrule}{0.4pt} 
    \fbox{%
        \includegraphics[width=0.85\columnwidth]{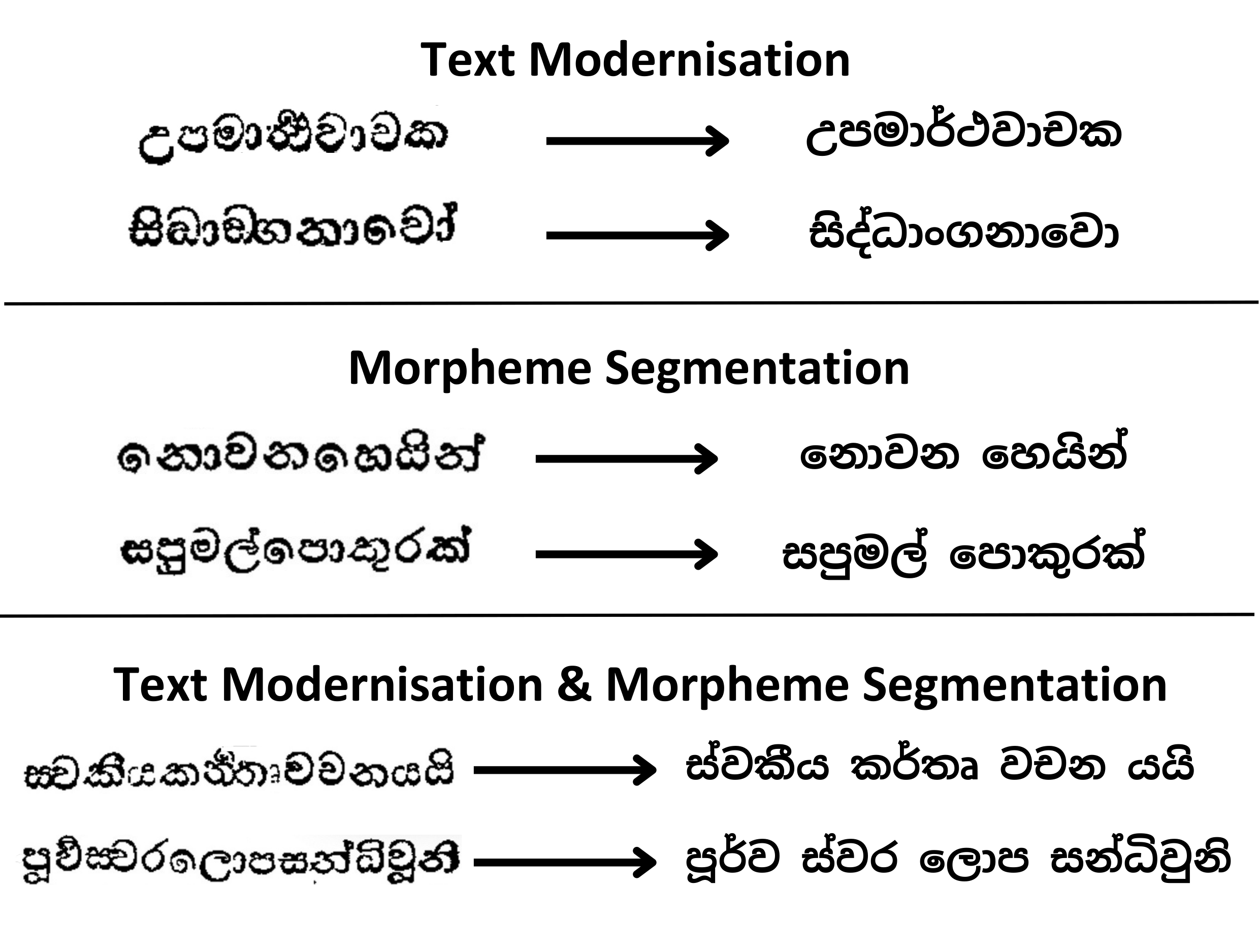}%
    }%
    \caption{Examples of Sinhala Text Modernisation and Morpheme Segmentation in \texttt{Document AI}.}
    \label{fig:doc_ai_ocr}
\end{figure}

\citet{jayatilleke-de-silva-2025-zero}, earlier in the process of creating \OldDataSetName, noted that both \texttt{Surya}\footL[suryaocr]{https://github.com/VikParuchuri/surya} and \texttt{Document AI}\footnote{\scriptsize \urlstyle{tt}\url{https://cloud.google.com/document-ai/}} were the best OCR engines in general. However, later in the process, under realistic conditions,~\citet{jayatilleke2025sidiac} observed that \texttt{Document AI} demonstrated superior performance compared to \texttt{Surya}, especially supported by the text modernisation and morpheme segmentation capabilities of the former in processing historical Sinhala as shown in Figure~\ref{fig:doc_ai_ocr}. Consequently, in this version of the corpus, we have chosen to use \texttt{Document AI} as the OCR engine for text extraction.

The methodology used for text extraction with \texttt{Document AI} closely followed the approach outlined in \OldDataSetName{} by~\citet{jayatilleke2025sidiac}. \texttt{Document AI} is a service offered by \texttt{Google Cloud Platform (GCP)}\footnote{\label{note:gcp}\scriptsize \urlstyle{tt}\url{https://cloud.google.com/}}. On this platform, we created a processor and utilised its API key to perform OCR. The processor can handle a maximum of 15 pages at a time; this limitation was also considered when determining our threshold for open-access books in the digital library of \texttt{Natlib}, as discussed in subsection~\ref{subsec:data_prep}. Throughout the process, we ensured that we obtained the model confidence for each page of every processed document. We then calculated the average confidence score, which is included in the metadata file of each book folder as discussed in subsection~\ref{subsec:metadata_creation}.

\subsection{Post-processing Extracted Text}

In this section, we discuss the manual post-processing of text extracted from documents using OCR technology. It was evident that \OldDataSetName{} effectively corrected several formatting issues during their post-processing stage, as discussed in the subsection~\ref{subsec:sidiac_lit}. 
However, there are still many other issues present in this corpus which we identified and discussed in the subsection~\ref{subsec:limits_sidiac} that need to be corrected.

\subsubsection{Handling Code-Mixing in Corpus}
\label{subsec:code_mix}

As noted by~\citet{jayatilleke2025sidiac}, the \OldDataSetName{} dataset contains content from Sanskrit, Pali, and English mixed with Sinhala. In this study, we carefully identified and removed all non-Sinhala content from the documents through manual inspection conducted by the authors, who are native Sinhala speakers. The removal of English content was straightforward, as there was very little of it in the entire corpus, and it was written in the Latin alphabet with no need to handle transliteration~\cite{de-mel-etal-2025-sinhala}. 

As observed by~\citet{gair1996sinhala}, the Sinhala script is also used for writing Pali and Sanskrit literature in Sri Lanka. But, there are no existing \texttt{LangID} models or even datasets trained to distinguish between the three languages written in the Sinhala script~\cite{de2025survey}.
Therefore, the authors differentiated between Sinhala and non-Sinhala texts using a simple visual negation rule, leading to the removal of 20 non-Sinhala books. These texts mainly contained Pali and Sanskrit \textit{gāthā}\footnote{\scriptsize A \textit{gāthā} is a metered verse or stanza within a Buddhist scripture.}, \textit{śloka}\footnote{\scriptsize A \textit{śloka} is a 32-syllable verse used in the many works of classical Sanskrit literature.}  and \textit{sūtra}\footnote{\scriptsize A \textit{sūtra} is a sacred text in Buddhism, comprising a large part of the Buddhist canon.}, with Sinhala explanations and commentaries. This was handled carefully, especially when word-for-word translations were presented in a single block. Such presentations made classification challenging due to the fact that Sinhala has loanwords directly borrowed from these languages that are used unchanged (\hspace*{-2pt}\raisebox{-0.5ex}{ 
    \includegraphics[height=1.5\fontcharht\font`\A]{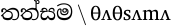} 
}\hspace{-2pt}), as well as terms that have been modified after borrowing but stopping short of being a full calque (\hspace*{-2pt}\raisebox{-0.5ex}{ 
    \includegraphics[height=1.5\fontcharht\font`\A]{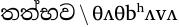} 
}\hspace{-2pt})~\cite{wijesiri2014building}.


\subsubsection{Inclusion of Special Tokens}
\begin{figure}[!htbp]
    \centering
    \setlength{\fboxsep}{1pt} 
    \setlength{\fboxrule}{0.4pt} 
    \fbox{%
        \includegraphics[width=0.95\columnwidth]{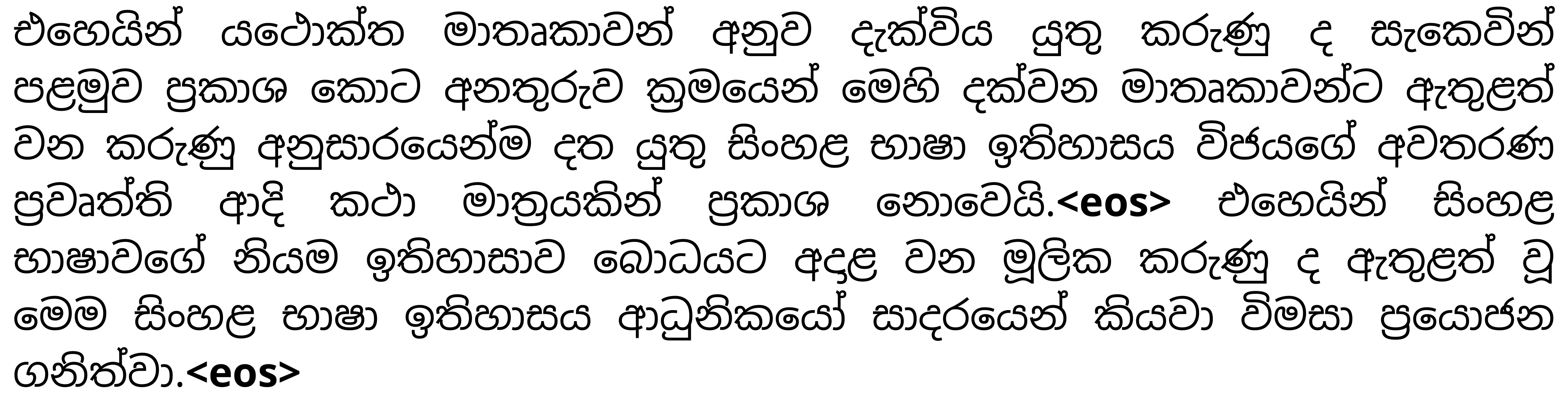}%
    }%
    \caption{An example block of text with \texttt{<eos>} tokens from "\textit{Sinhala Bhasha Ithihasaya}"}
    \label{fig:eos}
\end{figure}

\texttt{CCOHA}~\cite{alatrash2020ccoha} utilised a range of special tokens to convey different types of information. One token that stood out to us was the \underline{e}nd-\underline{o}f-\underline{s}entence (\texttt{<eos>}) indicator. This feature enables us to segment the entire corpus into individual sentences, thereby facilitating sentence-level diachronic analysis. Consequently, we adopted this method and inserted "\texttt{<eos>}" tokens, based on manual inspections, at the end of all sentences in our corpus. An example block of text with \texttt{<eos>} tokens is shown in Figure~\ref{fig:eos}. This manual check is important as the only native Sinhala punctuation mark, \raisebox{-0.1ex}{\includegraphics[height=1ex]{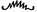}}(\raisebox{-0.3ex}{\includegraphics[height=2ex]{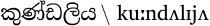}}), indicates the end of a text or a full section rather than a single sentence~\cite{jayatilleke2025sidiac}. All other punctuation used in Sinhala for contemporary or otherwise text was borrowed from Western languages during the colonial period (1505-1948)~\cite{kachru1994english}.

\begin{figure}[!htbp]
    \centering
    \setlength{\fboxsep}{1pt} 
    \setlength{\fboxrule}{0.4pt} 
    \fbox{%
        \includegraphics[width=0.95\columnwidth]{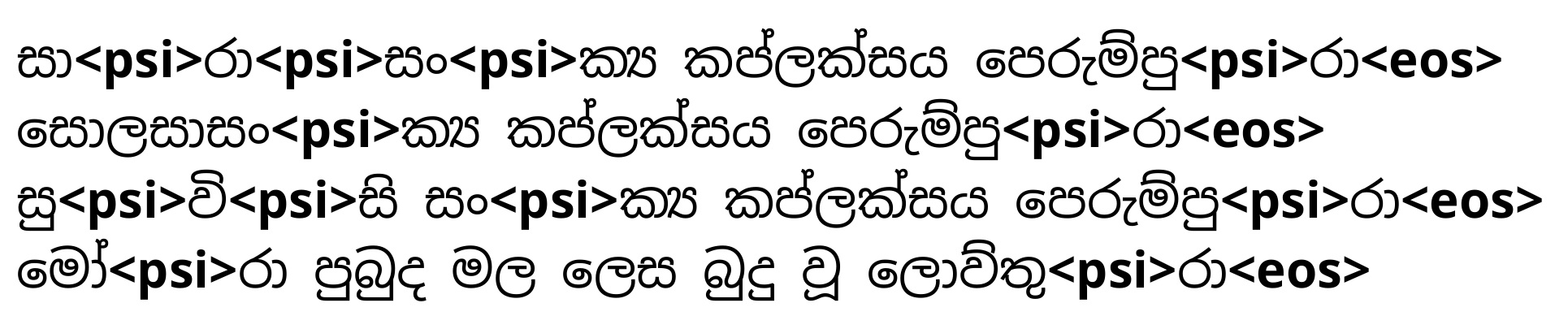}%
    }%
    \caption{An example poem with \texttt{<psi>} tokens from "\textit{Yasodharaawatha}". Note that we have added \texttt{<eos>} tokens at the end of each poem line.}
    \label{fig:psi}
\end{figure}

It is clear that \DataSetName{} includes poetry books, and the issue of poetry suffixes was discussed in section~\ref{subsec:limits_sidiac}. Fundamentally, what this means is that some Sinhala words appearing in poems have themselves split (sometimes at multiple places) using inserted spaces to better illustrate the rhyming pattern. We wanted to make sure, on one hand, that the words are properly formed for future word-level studies, and on the other hand, that this information on splitting present in the original texts is not lost. 
To facilitate this, we included a special \underline{p}oetry \underline{s}uffix-shift \underline{i}ndicator (\texttt{<psi>}) token, as shown in Figure~\ref{fig:psi}. Note that the inclusion \texttt{<psi>} token was conducted only when words were split, resulting in isolated characters, not when stand-alone words were shifted in a similar manner to illustrate the rhyming patterns. Therefore, any subsequent study that only concerns itself with word-level analysis and not how the original texts were presented may now do so by simply replacing the \texttt{<psi>} token with an empty string. 

\subsubsection{Corrections of Malformed Tokens}

%
The correction process for malformed tokens was extremely challenging and time-consuming, as we had to examine the entire corpus character by character.
During this step, we also addressed two formatting errors identified by~\citet{jayatilleke2025sidiac} which were closely related to the correction process: misplaced phrases and spacing issues. Tackling these formatting problems simultaneously with the character corrections was more efficient than correcting them separately, as it would have been significantly more time-consuming to traverse the entire corpus multiple times.

\subsubsection{Other Post-Processing Steps}
\label{subsec:other_post_steps}

The formatting errors noted by \OldDataSetName{} have also been corrected in this corpus. Page numbers and seals were removed, and indentation was standardised to the left, as precise indentation does not add significant value. Other issues, such as word and character spacing errors and misplaced text, were also addressed.
We identified four additional post-processing steps based on the limitations discussed in subsection~\ref{subsec:limits_sidiac}, which we took into account during the procedure.

\vspace*{-3.5mm}
\paragraph{Commentary Books:} This issue was partially addressed when we tackled the code-mixing problem, as most of the original content in the books was primarily in Pali and Sanskrit, with commentaries in Sinhala. For example, the series "\textit{Wishudhdhi Margaya - Dhwitheeya Baagaya}" was authored by "\textit{Buddhaghosa Himi}" in Pali during the 5th century, but we anchored its written date to the 13th century to match the Sinhala commentaries, as noted by~\citet{Sannasgala_2009}. In cases like "\textit{Sanna sahitha Salalihini Sandheshaya}," where both the original and commentary are in Sinhala, we used the original text's date due to the author's anonymity and the varied eras of the commentary writers.


\vspace*{-3.5mm}
\paragraph{Page Titles \& Content Tables:} We removed these during manual inspections because they provided contextually irrelevant information. Specifically, the different tables contained information that was difficult to represent effectively, ensuring contextual relevance.

\vspace*{-3.5mm}
\paragraph{Multi-Column Text:} We ensured that the multi-column text is formatted from top to bottom. This means that the content on the left side appears first, followed by the content on the right side, all within a single column. This approach standardises the layout of this content to match the structure of other text files.

\vspace*{-3.5mm}
\paragraph{Footnotes:} While the information in the footer often relates to the main content, it disrupts the flow within the text files of the books. Therefore, we decided to remove it to prioritise information fluency.




\subsection{Creation of Metadata Files}
\label{subsec:metadata_creation}


This step was based on \OldDataSetName{}, which holds comprehensive metadata for a diachronic corpus. Fields in these metadata files are derived from the literature.
Each folder, named after a book, contains a text file with the book's textual data and a \texttt{JSON} file with metadata~\cite{jayatilleke2025sidiac}, as shown in Table~\ref{tab:metaData}.

\begin{table}[h!tb]
    \centering
    \resizebox{0.48\textwidth}{!}{
    \begin{tabular}{l|c}
       \textbf{Metadata Tag}   & \textbf{Example} \\
       \hline
        \texttt{title}~\cite{he-etal-2014-construction}  &  \raisebox{-0.6ex}{\includegraphics[height=2ex]{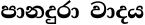}} \\
        \texttt{title\_en}~\cite{hellwig-etal-2020-treebank}$\bigtriangleup$  &  \textbf{Paanadhuraa Waadhaya}\\
        \texttt{author}~\cite{he-etal-2014-construction}$\diamond$  & \raisebox{-0.2ex}{\includegraphics[height=2.2ex]{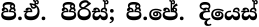}} \\
        \texttt{author\_en}~\cite{hellwig-etal-2020-treebank}$\bigtriangleup\diamond$  & \textbf{P.A. Peris; P.J. Dhiyes} \\
        \texttt{genre}~\cite{rognvaldsson-etal-2012-icelandic,kuvcera2015diakorp}  & \textbf{Non-Fiction; Religious} \\
        \texttt{issued\_date}~\cite{bouma-etal-2020-edges}$\star$  &  \textbf{1903}\\
        \texttt{written\_date}~\cite{mcgillivray2013tools}$\dagger$  & \textbf{1873} \\
        \texttt{ocr\_confidence}~\cite{jayatilleke2025sidiac}  &  \textbf{0.9945}\\
    \end{tabular}
    }
    \caption[]{Metadata records in \DataSetName{}.\\\tiny{
    $\bigtriangleup$ Transliterated (Romanised) by Native Sinhala speakers \\
    $\diamond$ If the authors were unknown, they were labelled as \textit{Unknown}\\
    $\star$ Always included as listed in the digital repository of \texttt{Natlib}\\
    $\dagger$ Included when applicable sourced from~\citet{Sannasgala_2009, Soratha_Godage}}}
    \label{tab:metaData}
\end{table}




The book genres were chosen based on details from~\citet{Sannasgala_2009} and evaluations by authors of this study who are native Sinhala speakers. The classification\footnote{\scriptsize Table~\ref{tab:meta_sidiac} in Appendix~\ref{app:eval_sdc} shows the classification of \WrittenDateBookCount{} books after the filtering process described in Section~\ref{sec:eval}.} occurs at two levels: the primary level divides books into `Fiction' and `Non-Fiction,' while the secondary level further categorises them into five classes: religious, history, poetry, language, and medical. 
It is important to note that the first level was applied to all documents, while the second level was applied to the books that fall into the selected specific categories. These genre classification levels are independent, and the frequency distributions of the genres across the two categories are shown in Tables~\ref{tab:freq_genre_1} and \ref{tab:freq_genre_2} in Appendix~\ref{app:genre_analy}.

\section{Evaluation of \DataSetName}
\label{sec:eval}

\begin{table*}[!htb]
    \centering
    \resizebox{0.99\textwidth}{!}{
    \begin{tabular}{l|c|c|r|r|r|r}
    \textbf{Corpus} & \textbf{Post-Processing} & \textbf{Date-based Filtering} & \tabh{c|}{Documents} & \tabh{c|}{Total Words} & \tabh{c|}{Unique Words} & \tabh{c}{Total Sentences} \\
    \hline
    \multirow{2}{*}{\OldDataSetName} & V.1.0 & \tick & \num{\OldBookCount} & \num{\OldWordCount} & \num{22837} & - \\
     & V.2.0 & \tick & \num{40} & \num{\OldFilteredWordCount} & \num{16025} & \num{2970} \\
     \hline    
    \multirow{2}{*}{\underline{\DataSetName{}}} & V.2.0 & \cross & \num{\BookCount} & \num{\WordCount} & \num{\UniqWordCount} & \num{11806} \\
     & V.2.0 & \tick & \WrittenDateBookCount & \num{\FilteredWordCount} & \num{21776} & \num{4363} \\
     \hline
    \end{tabular}}

    \caption{Summary of Information in \OldDataSetName{} and \DataSetName.}
    \label{tab:corpora_analysis}
\end{table*}

\DataSetName{}, contains a total of \num{\WordCount} words. This total was derived through regex-based filtering to isolate Sinhala text while excluding punctuation and Latin characters, followed by whitespace word tokenisation. In this corpus, there are \num{\UniqWordCount} unique Sinhala tokens, which represent \num{\UniqWordPerCen}\% of all words. 

Additionally, the portion of the dataset that includes written dates is composed of \num{\FilteredWordCount} words. This can be compared to the existing Sinhala diachronic corpus, \OldDataSetName, which initially claims to have \num{\OldWordCount} words. However, after following the same post-processing regimen in this study, it was found to contain only \num{\OldFilteredWordCount} words. The decrease in the number of tokens in \OldDataSetName{} occurred because five out of \num{\OldBookCount} literary works they use were removed from the corpus because they were entirely non-Sinhala. Some textual content in other books included in \OldDataSetName{} was also reduced due to the removal of non-Sinhala text. 

Out of the \num{\BookCount} books in the \DataSetName{} corpus, 135 are classified as Non-Fiction, while 50 are classified as Fiction. 
According to the secondary genre classification, 86 books are Religious, 54 are Poetry, 18 are History, 17 are Language, 5 are Medical, and 5 are Unclassified, with religious texts and poetry being the most prevalent. This predominance can be attributed to the close relationship between Sinhala literary culture and Theravada Buddhism, which has provided both the subjects and a framework for preserving texts. Furthermore, the influence of Sanskrit \textit{kavya} traditions and courtly patronage, which valued literary artistry and prestige, has also played a significant role~\cite{hallisey2003works, jayatilleke2025sidiac}. Genre analysis based on publication dates and written centuries is detailed in Appendix~\ref{app:genre_analy}.


A detailed comparison of \DataSetName, written date-based filtered \DataSetName, and the filtered \OldDataSetName{} is summarised in Table~\ref{tab:corpora_analysis}, focusing on the number of files, total and unique word counts, and total number of sentences.



Thereafter, we performed a Bag-of-Words analysis~\cite{davies2012expanding} per-century on the \DataSetName-\texttt{filtered}\footnote{\scriptsize \DataSetName{}-\texttt{filtered} is a subset of \DataSetName{} that has undergone both V.2.0 post-processing and date-based filtering, as per the information depicted in the final row of Table~\ref{tab:corpora_analysis}.} to examine the behaviour of neighbouring words associated with specific consistent words from the 13th to the 20th centuries.
In other words, this approach enables us to understand the diachronic changes in the contextual meanings of these selected consistent words. To carry out this procedure, we first partitioned the corpus based on the years when the documents were written, rounding up to the latest century in which each document could have been produced. Given that the 5th and 12th centuries have very few tokens, we decided to exclude them from this analysis because the number of common words identified would be significantly smaller if they were included. 

Next, after identifying common words from the 13th to the 20th centuries, we noted that many of these words were stopwords. We conducted a stopword analysis for the entire corpus using z-scores~\cite{jayatilleke2025sidiac, wijeratne2020sinhala}, calculating \( -\infty < Z < 5.293\) for a 99.80\% threshold. However, due to a corpus bias towards Buddhism and history, some words that should not be classified as stopwords appeared in the identified list, which we manually removed. After that, we eliminated stopwords from the list of century-specific consistent words. For the remaining words, we conducted the BoW analysis with a window span of ±10 following the methodology by~\citet{davies2012expanding}. Next, we removed stopwords from the identified neighbouring word sets as well. This is to ensure that the subsequent analysis includes only words that provide contextual meaning to these consistent words.

We identified \ConsistentKeys{} consistent words from the 13th to the 20th century. The limited number of consistent words is due to the lack of lemmatisation, as the Sinhala language does not have highly accurate morphological analysers~\cite{de2025survey}, particularly for historical Sinhala~\cite{jayatilleke2025sidiac}. From these consistent words, we selected two well-known terms that possess multiple meanings for further examination. We then identified their top 200 collocates based on their longevity across centuries, as detailed in Appendix~\ref{app:bow-analy}. Finally, we manually picked and examined the neighbouring words of these collocates to qualitatively analyse their semantic relationships.


The first example is the word "\raisebox{-0.5ex}{%
\includegraphics[height=1.5\fontcharht\font`\A]{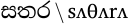}}," which has multiple senses; it can mean the number \textit{four}, \textit{skills}, or \textit{thief}, depending on the context. Additionally, the number four carries religious significance. As indicated in Table~\ref{tab:neigh_sath} in Appendix~\ref{app:bow-analy}, words related to \textit{learning} and \textit{education} are visible in the 13th, 16th, 19th, and 20th centuries, with a more distributed frequency over time. 
The only term associated with the meaning of \textit{thief} appears to be present solely in the 19th century and is noted for its very low frequency. The remaining words with the meanings related to \textit{wisdom} and \textit{knowledge}, \textit{direction}, \textit{value}, and \textit{mathematics} are all connected to the sense of "\raisebox{-0.5ex}{%
\includegraphics[height=1.5\fontcharht\font`\A]{Figures/si_003.pdf}}" as the number \textit{four}. Given the religious importance of this number, references to hell (found only in the 13th and 15th centuries) and references to knowledge and wisdom (scattered between the 15th and 19th centuries) likely pertain to the four types of wisdom and the four hells in Buddhism.

The second example is the word "\raisebox{-0.5ex}{%
\includegraphics[height=1.5\fontcharht\font`\A]{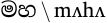}},"  which also has multiple senses. It can signify \textit{esteemed, great, or sacred}; \textit{powerful or strong}; and \textit{big, large, or massive}. As shown in Table~\ref{tab:neigh_maha} in Appendix~\ref{app:bow-analy}, words related to  \textit{great}, \textit{esteemed}, and \textit{sacred} were prevalent during the 13th and 14th centuries. However, their usage declined in the 15th through 19th centuries, before experiencing a resurgence in the 20th century. Words associated with the meanings of \textit{powerful} and \textit{strong} showed relatively low frequency during the 13th and 14th centuries. Their usage was entirely absent from the 15th to the 19th centuries, with no occurrences of associated words. In contrast, the 20th century saw a significant resurgence in the frequency of these words. Additionally, words related to the meanings of \textit{big}, \textit{large}, or \textit{massive} were dominant during the 13th to 15th centuries but appeared to disappear afterwards, only to re-emerge in the 20th century, where they collectively accounted for just two occurrences amongst all related words.

\section{Conclusion} 

\DataSetName{} is the largest Sinhala diachronic corpus to date, comprising \WordCount{} word tokens. After filtering for written dates, it contains \FilteredWordCount{} word tokens, making it a comprehensive resource for temporal linguistic studies. \DataSetName{} will serve as an important corpus for exploring lexical semantic drift by developing diachronic embeddings, tracking neologisms, examining grammatical changes, conducting historical language modelling, and supporting corpus-based lexicography.

The creation of this corpus involved a meticulous process, which included identifying literature from the \texttt{Natlib} of Sri Lanka through \OldDataSetName{}. This was followed by data filtering and preprocessing, written date annotation, text extraction from PDFs, and extensive post-processing. The carefully created metadata files include important information, specifically genre classifications. This classification enables synchronic linguistic studies to focus on specific time periods, adding significant value to the corpus.

The corpus spans the years from 1800 to 1955 CE based on publication dates and from the 5th century to the 20th century CE based on written dates. A thorough analysis of the complete corpus was conducted based on the metadata of all the books, with a comprehensive examination at the word token level. This analysis aimed to identify key findings within the corpus, which were then compared to the only other available Sinhala diachronic corpus, \OldDataSetName{}. 

\section{Limitations}

The creation of the \DataSetName{} corpus faced various limitations and challenges as listed below.

\paragraph{Written Date Annotation:} This process was carried out using the available information from the books by~\citet{Sannasgala_2009} and~\citet{Soratha_Godage}. However, we were only able to complete this for \WrittenDateBookCount{} out of \BookCount{} books. The task is extremely challenging, as identifying a specific written time period in historical documents is inherently difficult.

\paragraph{Commentary Books:}  
The identified books that primarily include the term `\raisebox{-0.35ex}{
    \includegraphics[height=1.5\fontcharht\font`\A]{Figures/si_052.pdf} 
}' (meaning commentaries) must feature two written dates: one for the original text and another for the commentary. Although this phenomenon was partially addressed, as mentioned in subsection~\ref{subsec:other_post_steps}, a comprehensive approach to dating these texts was not implemented. This is particularly important for identifying the time periods of both the original work and its commentaries, especially when multiple versions of the commentaries exist.

\paragraph{Lexical Annotation:} 
Most of the similar studies on high-resource languages~\cite{mcgillivray2013tools,rognvaldsson-etal-2012-icelandic,davies2012expanding}, as well as some low-resource languages~\cite{hellwig-etal-2020-treebank}
, have conducted PoS and other downstream annotations on their corpora. 
However, we were unable to perform similar lexical annotations for PoS tagging in the \DataSetName{} corpus. This limitation is primarily due to the lack of accurate Sinhala PoS taggers available for our use~\cite{de2025survey}. Consequently, this also precludes us from other downstream lexical annotations for which PoS is a prerequisite, such as dependency parsing~\cite{durrell2012germanc} and generating tree banks~\cite{hellwig-etal-2020-treebank}.
Moreover, while the text dating process was fully supported by existing resources, as discussed in subsection~\ref{subsec:writte_date}, the post-processing was carried out by the authors of this study. They followed an iteratively improving set of instructions while working on discrete subsets of documents. Consequently, conducting inter-annotator agreement was not feasible.

\section{Acknowledgements}

The creation of the \DataSetName{} corpus was made possible thanks to the valuable contributions of several individuals and organisations. We acknowledge the support from the \textit{LK Domain Registry} for funding granted to publish this paper. We sincerely thank \texttt{Padma Bandaranayake}, director of the \textit{National Library and Documentation Centre}, for her assistance with data acquisition during the development of both \OldDataSetName{} and \DataSetName{}. We appreciate the expertise of \texttt{Ven. Thimbiriwawa Sirisumana}, a senior lecturer at the \textit{Bhiksu University of Sri Lanka}, along with \texttt{Nalaka Jayasena} as Sinhala linguists, whose insights were crucial for validating our text dating and genre classification procedures.

\section{Bibliographical References}\label{sec:reference}
\bibliographystyle{lrec2026-natbib}
\bibliography{anthology-1,anthology-2,lrec2026-example}

\section{Language Resource References}
\label{lr:ref}
\bibliographystylelanguageresource{lrec2026-natbib}
\bibliographylanguageresource{languageresource}



\appendix

\section{Exisiting Diachronic Corpora}
\label{app:AllDS}

\begin{table*}[!htb]
    \centering
    \resizebox{!}{0.47\textheight}{
        \begin{tabular}{|l|l|r|c|c|}
        \hline
        \multirow{2}{*}{\textbf{Dataset}} & \multicolumn{2}{c|}{\textbf{Language}} & \multirow{2}{*}{\textbf{Time Span}} & \multirow{2}{*}{\textbf{Token Count}} \\
        \hhline{~--~}
         & \textbf{Name} & \textbf{Class} &  & \\
        \hline
        \DSTable
        \end{tabular}
        }
    \caption{Summary of the \LitCount{} diachronic corpora surveyed; including and sorted by the language class as defined by~\citet{ranathunga-de-silva-2022-languages}.}
    \label{tab:AllDS}
\end{table*}

\begin{figure*}[!htbp]
    \centering
        \includegraphics[width=\textwidth]{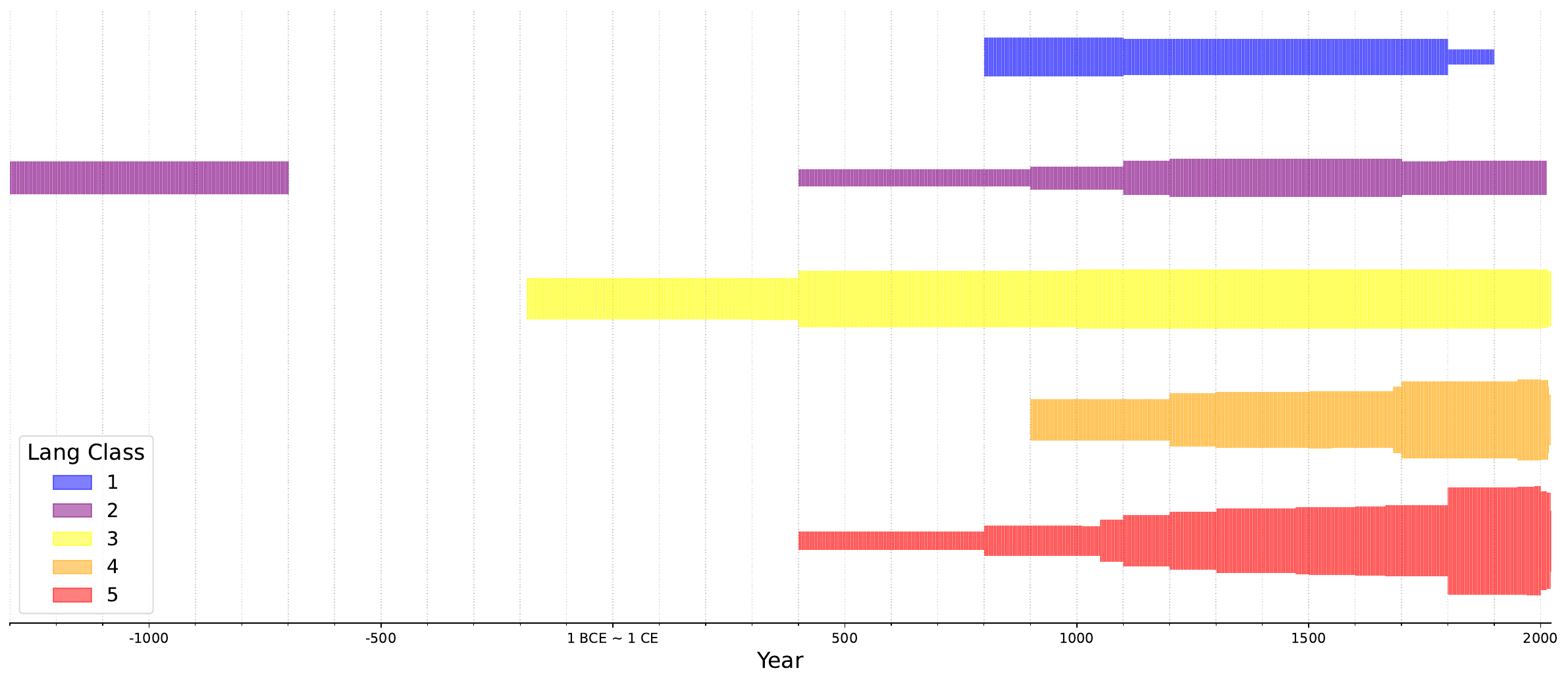}%
    \caption{Multi-class timeline of presence and intensity. The visualisation illustrates longitudinal shifts in language class activity, with vertical bar thickness serving as a proxy for rate magnitude.}
    \label{fig:rate_timeline}
\end{figure*}

Across the full spectrum of diachronic corpus initiatives, a set of cross-cutting design decisions becomes apparent: depth versus scale, philological precision versus computational efficiency, and standardisation versus historical authenticity. Diverse projects illustrate how corpus-building philosophies have evolved from manually curated, normalised text-bases toward dynamic, monitor-style resources that emphasise scalability and interoperability~\cite{davies2002corpus,zampieri2013colonia,vaamonde2015ps,wissik2018parlat}. These transitions map broader epistemological shifts within corpus linguistics and natural language processing (NLP): from human-led lexicography toward automated, model-ready datasets capable of serving more as computational benchmarks rather than traditional linguistic archives.

A central divergence among corpora concerns their treatment of orthographic and morphological variability. Early reference corpora~\cite{davies2002corpus,davies2009creating,onelli-etal-2006-diacoris} relied on strong normalisation pipelines that maximised searchability but at the expense of variant preservation. Later efforts~\cite{sanchez-marco-etal-2010-annotation,gelumbeckaite2012senosios,klein2016handbuch} adopt stand-off annotation that separates paleographic and normalised layers, allowing users to switch between authenticity and comparability. This shift may have been a recognition of the fact that orthographic variation is a proxy for diachronic phonology, sociolinguistic practice, and textual transmission rather than mere noise in the dataset. \citet{contreras2009hacia} and \citet{vaamonde2015ps} embed critical editions within digital environments to preserve variant readings and editorial commentary. Similarly, the Lithuanian corpus~\cite{gelumbeckaite2012senosios} uses parallel alignment to source languages, treating normalisation as an interpretive act rather than preprocessing.

On the matter of parallelism, parallel corpora introduced new possibilities for cross-temporal and cross-lingual anchoring. The \texttt{PROIEL}~\cite{haug2008creating} and \texttt{TOROT}~\cite{eckhoff2015linguistics} treebanks exploit Greek-Slavic alignment to model syntactic diffusion, while the \texttt{EDGeS} Bible corpus~\cite{bouma-etal-2020-edges} and Old Lithuanian \texttt{sLieKKas}~\cite{gelumbeckaite2012senosios} demonstrate translation-based calibration. Alignment also extends beyond text: embedding-based temporal alignment~\cite{steuer-etal-2024-modeling,gribomont-2023-diachronic-contextual} synchronises vector spaces across centuries, enabling semantic continuity analysis. 

Classical synchronic corpora~\cite{durrell2012germanc,geyken2011deutsche,klein2016handbuch} sought coverage across genres and dialects, whereas later diachronic designs pursued temporal comparability~\cite{kutuzov-pivovarova-2021-three,rodina-kutuzov-2020-rusemshift,chen-etal-2022-lexicon}. These later works adopt fixed vocabularies or rating scales that allow semantic trajectories to be traced over time without lexical drift in the sample. At the same time, domain-specific corpora such as the \textit{Royal Society Corpus}~\cite{kermes-etal-2016-royal} or the \textit{Austrian parliamentary corpus}~\cite{wissik2018parlat} redefine balance in sociological rather than statistical terms. 
The principle of temporal stratification is also visible in \citet{onelli-etal-2006-diacoris} and \citet{braun-2022-tracking}, where comparability is anchored by slicing data into socially coherent historical intervals.

Some~\cite{rognvaldsson-etal-2012-icelandic,lavrentiev2011syntactic,xue2005penn} integrate morphosyntactic detail while maintaining XML or CoNLL interoperability. Others~\cite{sagot-etal-2006-lefff,koprivova-etal-2014-mapping,favaro-etal-2022-towards} work on the basis of how dictionary alignment and linked data can feed into corpus annotation pipelines. Then, a pragmatic synthesis between linguistic accuracy and NLP scalability achieved through the combination of automatic parsing and manual curation can be seen in some later works~\cite{eide2016swedish,eythorsson2014greinir,gifu2016tracing}. However, even quite recently, work that builds on Universal Dependencies corpora~\cite{scannell-2022-diachronic} still prioritises consistent syntax over exhaustive morphology.

\texttt{ANNIS}-based ecosystems~\cite{klein2016handbuch,hellwig-etal-2020-treebank} and \texttt{CQPWeb} deployments~\cite{zampieri2013colonia,wissik2018parlat} coexist with emerging \texttt{RDF} and \texttt{Linked Data} graphs~\cite{armaselu-etal-2024-llodia}. Open pipelines such as \citet{erjavec2015imp} and \citet{sanchez2013open} provide transparent, reproducible workflows, while \citet{onelli-etal-2006-diacoris} and \citet{contreras2009hacia} show the persistence of regional, non-commercial academic hosting. The infrastructural contrast between reflects differing user communities: \textit{philologists favour manual curation, whereas computational researchers value standardised export formats}.

Diachronic corpora are becoming valuable benchmarks for evaluating NLP rather than just serving as simple repositories of data. The \texttt{RuSemShift}~\cite{rodina-kutuzov-2020-rusemshift} and \texttt{RuShiftEval}~\cite{kutuzov-pivovarova-2021-three} datasets quantify Russian lexical change through human judgments, while \citet{chen-etal-2022-lexicon} and \citet{cassese-etal-2025-diacu} generalise these methods to Chinese and Church Slavonic, respectively. Similarly, \citet{hellwig-etal-2020-treebank} and \citet{he-etal-2014-construction} link diachronic syntax to compositional semantics via treebank-driven modelling. These benchmarks complete what we observe as the constructive methodological loop: \textit{richly annotated corpora inspire evaluation datasets, which in turn motivate more sophisticated annotation schemes}.

In high-resource languages, having already established ample amounts of general domain corpora, diachronic inquiry has started to broaden (or rather \textit{focus}) into new domains. The \textit{Royal Society Corpus}~\cite{kermes-etal-2016-royal} and long-term scientific surprisal studies~\cite{steuer-etal-2024-modeling} capture historical change in scientific prose. Diachronic change is tracked in direct political registers~\cite{wissik2018parlat} as well as the resultant legal policy reform~\cite{braun-2022-tracking}. Media~\cite{braun-2022-tracking}, religious translation~\cite{gelumbeckaite2012senosios}, and even private correspondence~\cite{vaamonde2015ps} are tracked, diversifying textual evidence for everyday and liturgical language. 

Linguistic corpora now further intersect with cultural analytics and cognitive modelling. Hybrid projects such as \citet{o2024diachronic} combine embeddings with surprisal analysis to chart shifts in scientific discourse; \citet{gribomont-2023-diachronic-contextual} used transformer architectures to measure contextual semantic drift. \texttt{TEI}-encoded multimodal annotation in \citet{gelumbeckaite2012senosios} and image-aligned archives in \citet{vaamonde2015ps} bridge the continuity between philological and computational paradigms. Cross-linkages between \citet{bouma-etal-2020-edges} and \citet{armaselu-etal-2024-llodia} further reveal a convergence between diachronic linguistics and the semantic web.

To visualise the evolution of language resources, we calculate the Annualised Contribution Rate for each diachronic corpus. By aggregating these annual contributions based on the resource categories proposed by~\citet{joshi-etal-2020-state}, we were able to capture the yearly rate of data availability for each category with a uniform chronological distribution assumption applied to each corpus. Since the raw data volumes for high-resource languages are millions of times larger than those for low-resource languages, we applied a base-10 logarithmic transformation to the results. Finally, we cumulatively aggregated these values to visualise the diachronic progression of language resources, illustrating how the proportions of data availability shift across different resource classes over time, as depicted in Figure~\ref{fig:rate_timeline}.

When we analyse the yearly rate of data availability for category 1 languages, it becomes evident that data availability begins around 800 CE with the introduction of Slavonic corpora~\cite{cassese-etal-2025-diacu,eckhoff2015linguistics,haug2008creating}. A slight decline is observed around 1100 CE, attributed to the completion of the \textit{Old Church Slavonic} subsets of the \texttt{TOROT}~\cite{eckhoff2015linguistics} and \texttt{PROIEL}~\cite{haug2008creating} corpora. This decrease is followed by a minor resurgence in 1114 CE with the initiation of DACON~\cite{o2024diachronic}. Another small decline occurs in 1251 CE, linked to the conclusion of the \textit{Kiev-era Old East Slavic} subset of \texttt{TOROT}~\cite{eckhoff2015linguistics}. A particularly significant drop in data availability is observed in 1800 CE, resulting from the completion of the \textit{Old Church Slavonic} subset of \texttt{DIACU}~\cite{cassese-etal-2025-diacu}, which ultimately leads to a complete absence of data beyond 1900 CE. Similarly, when we analyse category 2 languages from 1300 BCE to 699 BCE, we notice a consistent rate of data availability influenced by \texttt{DCS}~\cite{hellwig-etal-2020-treebank}. After a complete absence of data in 699 BCE, there is a resurgence in 400 CE with the introduction of \OldDataSetName{}~\cite{jayatilleke2025sidiac}. Multiple spikes in data availability occur until it peaks at 1600 CE, influenced by \texttt{Pre-Standard Irish}~\cite{scannell-2022-diachronic}, \texttt{Greinir skáldskapar}~\cite{eythorsson2014greinir}, \texttt{Menota-Icelandic}~\cite{haugen2008menota}, and \texttt{IcePaHC}~\cite{rognvaldsson-etal-2012-icelandic}. Following this peak, the rate fluctuates with declines until 2013 CE, ultimately leading to a complete absence of data with the conclusion of \texttt{IcePaHC}.

For category 3 languages, data begins to appear around 186 BCE with the introduction of \texttt{LatinISE}~\cite{mcgillivray2013tools}. It shows consistent positive growth until 400 CE, when a significant spike occurs with the inception of \texttt{GNC}~\cite{gippert2015structuring} and the \textit{Armenian} subset of \texttt{PROEIL}~\cite{haug2008creating}. Following this, the data fluctuates around a similar rate until 2016 CE, when a sudden drop is observed due to the conclusion of \texttt{GNC}~\cite{gippert2015structuring}. Ultimately, the data reaches a complete absence in 2024 CE with the end of the \textit{Hebrew} subset of \texttt{LLODIA}~\cite{armaselu-etal-2024-llodia}. Furthermore, for category 4 languages, data starts to emerge around 900 CE with the introduction of \texttt{HGDS}~\cite{simon2014corpus}. There is a noticeable spike in 1200 CE, influenced by the inception of the \textit{Norwegian} subset of \texttt{Menota}~\cite{haugen2008menota}, and \texttt{MAþiR}~\cite{text2024MApiR}. This trend continues with fluctuations until 1700 CE, when another steep spike occurs due to \texttt{RuShiftEval}~\cite{kutuzov-pivovarova-2021-three} and \texttt{Kubhist}~\cite{lilljegren2018introduktion}. The data maintains a pattern of stable fluctuations until 2015 CE. From 2015 CE onward, the rate gradually declines due to the conclusions of \texttt{The Swedish Culturomics Gigaword corpus}~\cite{eide2016swedish}, \texttt{RuShiftEval}~\cite{kutuzov-pivovarova-2021-three}, and \texttt{RuSemShift}~\cite{rodina-kutuzov-2020-rusemshift}, ultimately reaching a complete absence of data by 2021 CE with the completions of the \textit{Swedish} and \textit{Dutch} subsets of \texttt{EDGeS}~\cite{bouma-etal-2020-edges}.

Finally, for category 5 languages, data begins to emerge around 400 CE with the introduction of the \textit{English} subset of the \texttt{ISWOC}. The rates demonstrate a steady increase until 1200 CE. From 1200 CE to 1721 CE, growth continues due to the inclusion of multiple corpora, as indicated in Table~\ref{tab:corpora_analysis}. A significant spike occurs in 1800 CE with the introduction of the \texttt{ENGALL}~\cite{davies2012expanding}, \texttt{ENGFIC}~\cite{davies2012expanding}, \texttt{GERALL}~\cite{schneider1998adding}, and \texttt{FREALL}~\cite{sagot-etal-2006-lefff} corpora. This trend persists, showing only minor fluctuations until 2020 CE, which marks the conclusion of the \texttt{German Court Decisions}~\cite{braun-2022-tracking}, as well as the \textit{English} and \textit{German} subsets of \texttt{EDGeS}~\cite{bouma-etal-2020-edges}. As of 2023 CE, there is a complete absence of data following the conclusion of \texttt{EZLN}~\cite{gribomont-2023-diachronic-contextual}.

Recent advances highlight low-resource diachronic NLP as a central frontier. The resource constraint is mitigated using tailored annotation for Icelandic~\cite{rognvaldsson-etal-2012-icelandic,eythorsson2014greinir} and Irish~\cite{scannell-2022-diachronic}. Treebank initiatives adapt dependency parsing to Sanskrit~\cite{hellwig-etal-2020-treebank} and Slavic languages~\cite{haug2008creating,eckhoff2015linguistics}, while \citet{jayatilleke2025sidiac} and \citet{cassese-etal-2025-diacu} extend these frameworks to South Asian and Church Slavonic corpora, respectively.  

\section{Limitations of \OldDataSetName{}}
\label{app:limits_sidiac}

\begin{figure*}[!htbp]
    \centering
    \setlength{\fboxsep}{1pt} 
    \setlength{\fboxrule}{0.4pt} 
    \fbox{
        \includegraphics[width=0.98\textwidth]{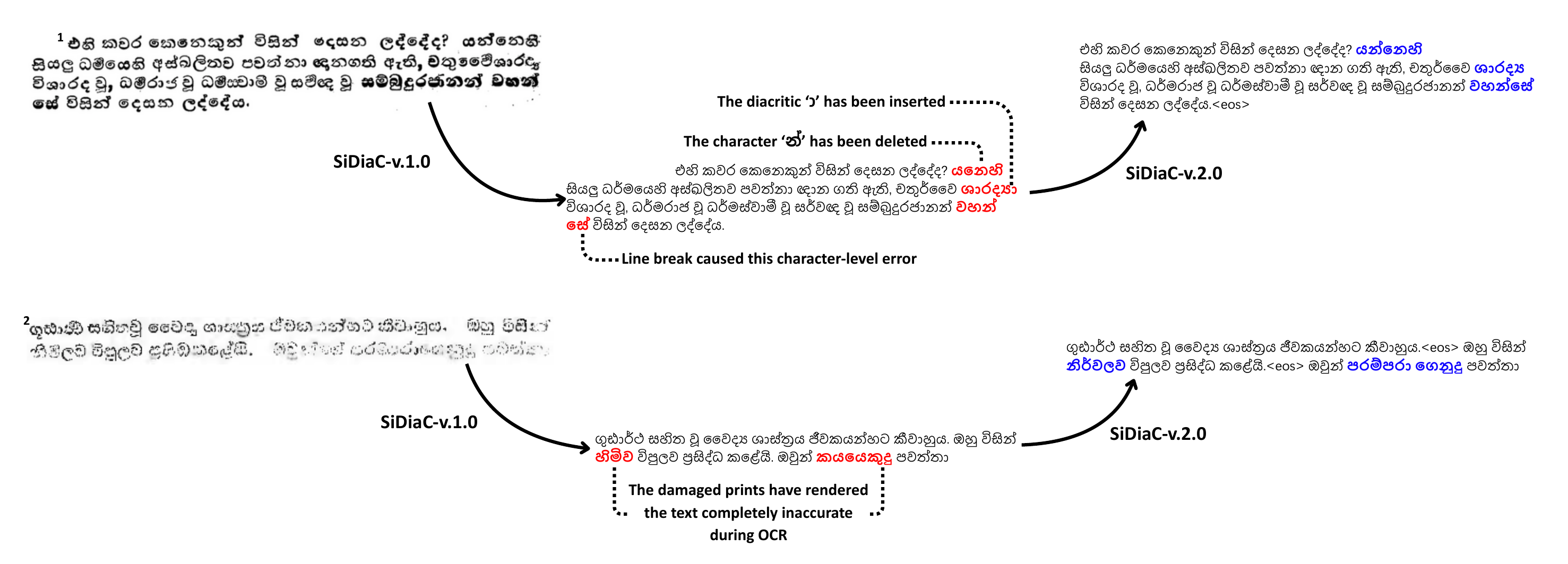}%
        }
    \caption{Examples of malformed tokens in \OldDataSetName{} and the way \DataSetName{} has corrected them. These sample images were taken from the books $^1$\textit{Anagathawanshaya: Methe Budu Siritha} and $^2$\textit{Sarartha Sangrahawa: Prathama Bhagaya}.}
    \label{fig:malformed-tok}
\end{figure*}

During the post-processing of \OldDataSetName, only five formatting issues were addressed, as detailed in section~\ref{subsec:limits_sidiac}. However, the most significant potential errors related to word or character identification from the OCR process were not addressed. These errors can be identified as spell-correction domain errors~\cite{gunathilake2025lmspell}, which involve substitutions, deletions, and insertions of characters and words. Such mistakes primarily occurred due to noise in the scanned documents and visible damage, as these documents are decades or even centuries old. These possibly led to misinterpretations by the OCR engine. Additionally, line breaks that caused words to split were also examined during the post-processing of \DataSetName{}. A few examples of these issues are included in Figure~\ref{fig:malformed-tok}.

\begin{figure}[!htbp]
    \centering
    \setlength{\fboxsep}{1pt} 
    \setlength{\fboxrule}{0.4pt} 
    \fbox{%
        \includegraphics[width=0.98\columnwidth]{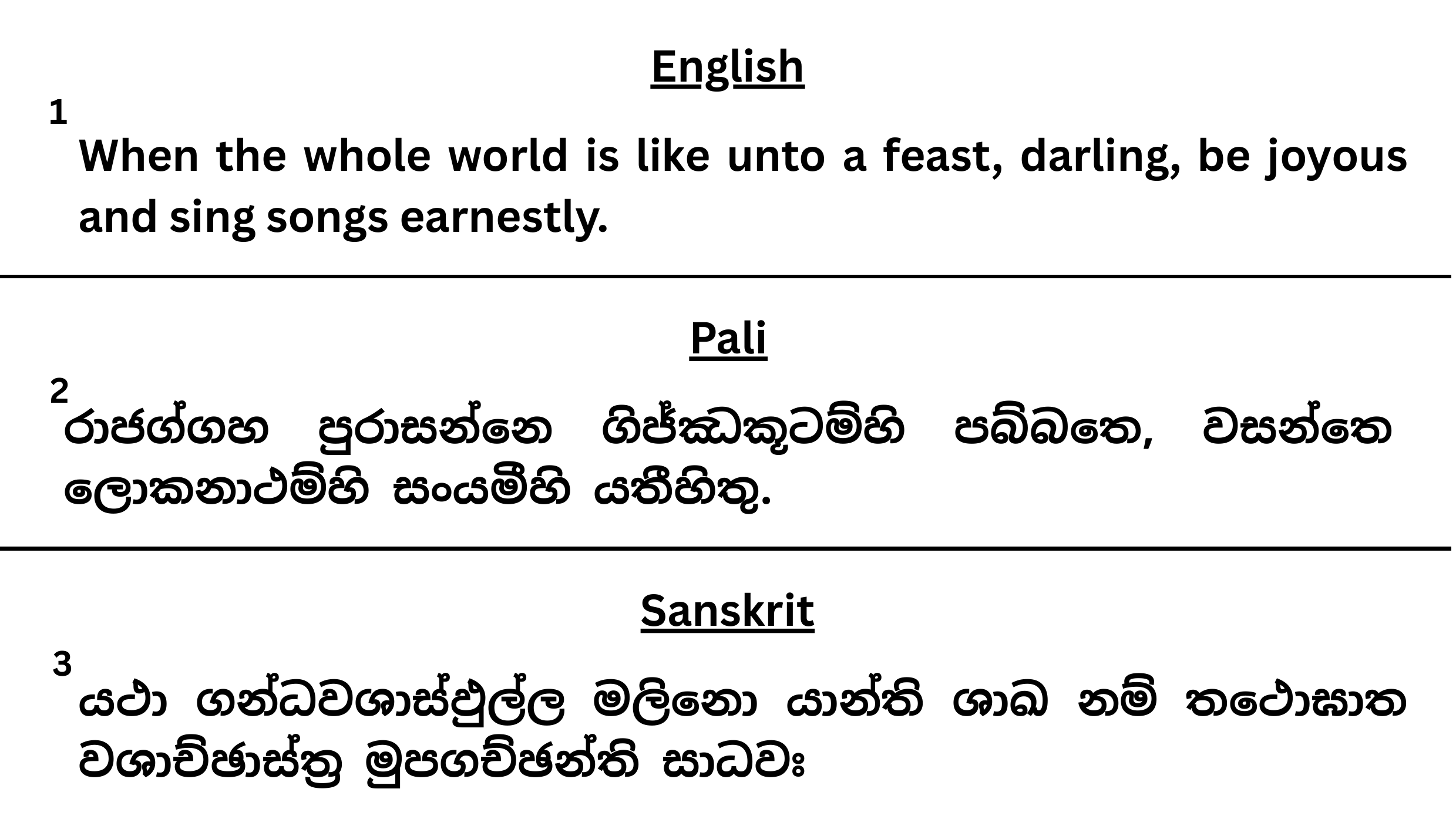}%
    }%
    \caption{Examples of code-mixed content in \OldDataSetName{}. These sample texts were taken from the following books: $^1$\textit{Jubili Warnanawa} - English in Latin script, $^2$\textit{Adhimasa Dheepanaya} - Pali in Sinhala script, $^3$\textit{Moggallana Panchika Pradeepaya} - Sanskrit in Sinhala script.}
    \label{fig:code-mixed}
\end{figure}

It was observed that \OldDataSetName{} includes a mixture of Pali written in Sinhala script, Sanskrit in Sinhala script, and English in Latin script, as illustrated in Figure~\ref{fig:code-mixed}. To ensure that \DataSetName{} focuses solely on Sinhala, we removed texts from these three languages using the negation rule discussed in subsection~\ref{subsec:code_mix}.

\begin{figure}[!htbp]
    \centering
    \setlength{\fboxsep}{1pt} 
    \setlength{\fboxrule}{0.4pt} 
    \fbox{%
        \includegraphics[width=0.98\columnwidth]{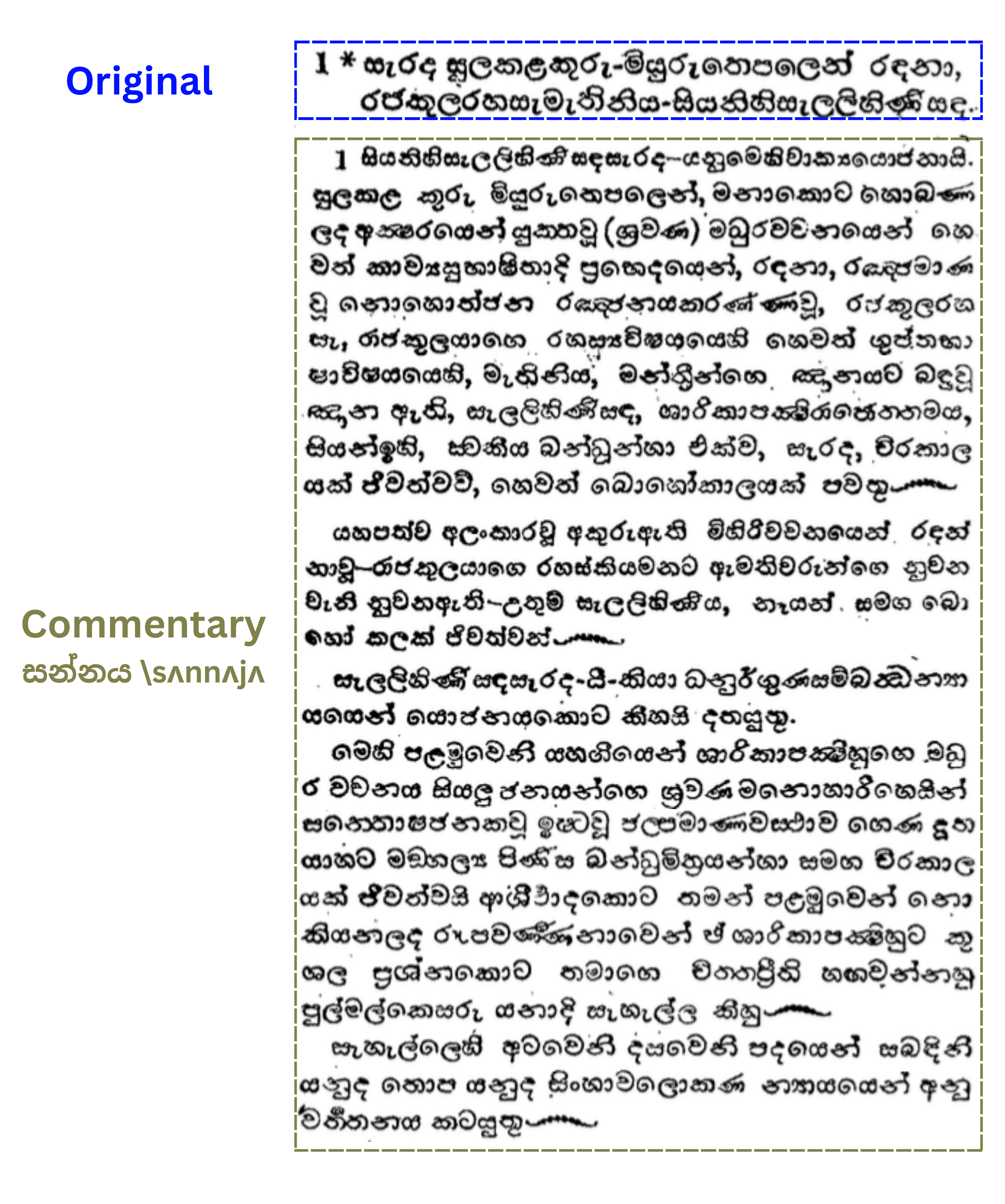}%
    }%
    \caption{An example of commentaries in \OldDataSetName{} from "\textit{Sanna sahitha Salalihini Sandheshaya}".}
    \label{fig:commentaries}
\end{figure}

The books that have titles including the word "\raisebox{-0.35ex}{%
     \includegraphics[height=1.5\fontcharht\font`\A]{Figures/si_052.pdf}%
}" are commentaries on earlier works. Additionally, there are some books, such as "\textit{Wishudhdhi Margaya - Dhwitheeya Baagaya}" that do not contain this keyword but also provide commentaries. Consequently, the content of these texts may span two different time periods, featuring text from the original work followed by the commentary. An example of this phenomenon can be found in Figure~\ref{fig:commentaries}. Furthermore, the way in which the \DataSetName{} post-processing addressed this issue is discussed in subsection~\ref{subsec:other_post_steps}.

\begin{figure}[!htbp]
    \centering
    \setlength{\fboxsep}{1pt} 
    \setlength{\fboxrule}{0.4pt} 
    \fbox{%
        \includegraphics[width=0.98\columnwidth]{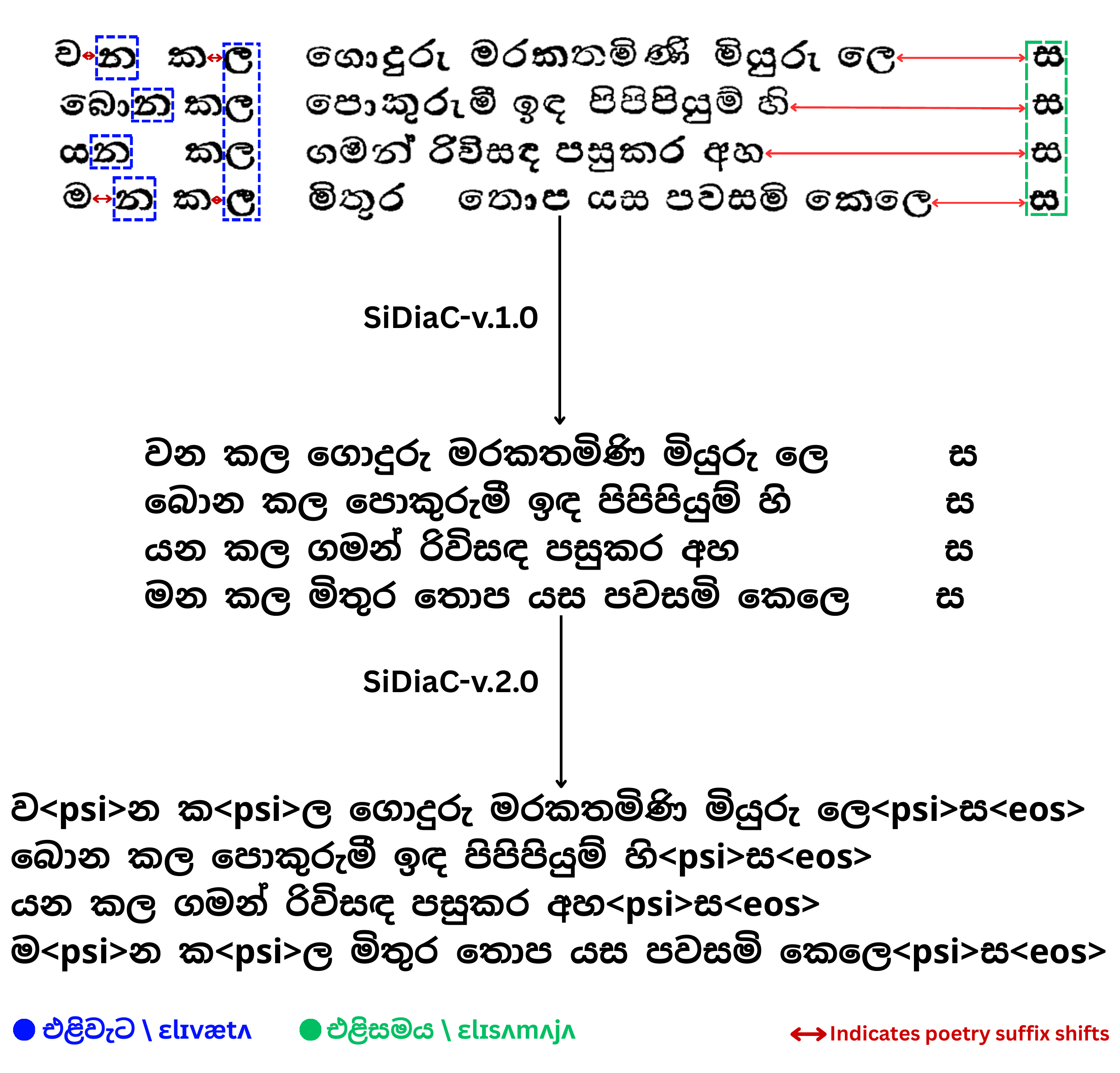}%
    }%
    \caption{An example of poetic suffix shifts from "\textit{Hansa Sandheshaya}" and the way \OldDataSetName{} and \DataSetName{} handled them.}
    \label{fig:poetry_suff}
\end{figure}

The poems in both \OldDataSetName{} and \DataSetName{} emphasize rhyming through the use of \hspace*{-3pt}\raisebox{-0.5ex}{
    \includegraphics[height=1.45\fontcharht\font`\A]{Figures/si_046.pdf} 
} to ensure that the rhyme sound at the end of each line, which typically stands out from the final word, while \raisebox{-0.5ex}{
    \includegraphics[height=1.45\fontcharht\font`\A]{Figures/si_047.pdf} 
} underscores this separated sound through repetition, isolating or detaching the rhyming sound from the complete word at the beginning and middle, as illustrated in Figure~\ref{fig:poetry_suff}.

\begin{figure}[!htbp]
    \centering
    \setlength{\fboxsep}{1pt} 
    \setlength{\fboxrule}{0.4pt} 
    \fbox{%
        \includegraphics[width=0.98\columnwidth]{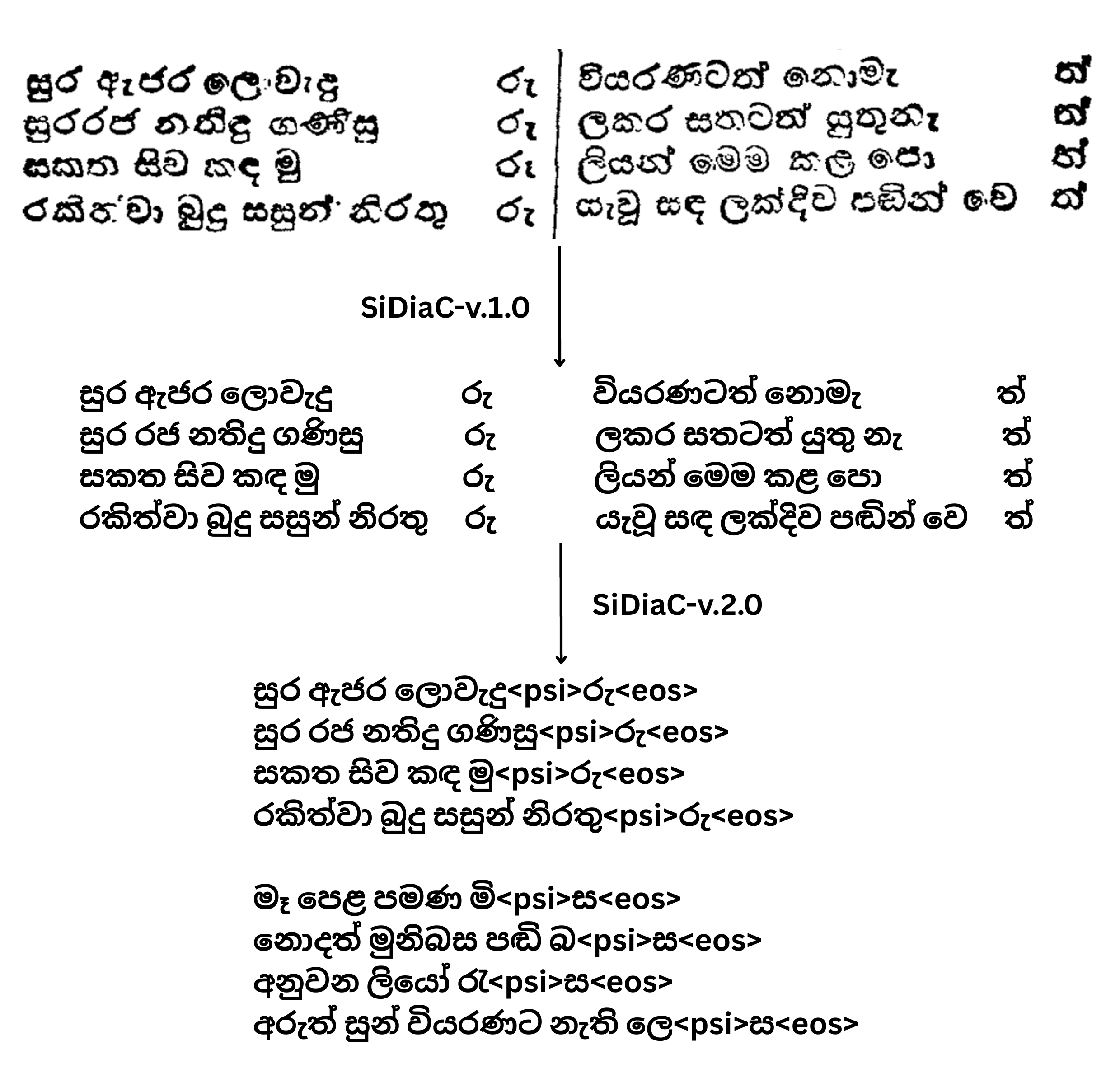}%
    }%
    \caption{An example of multi-column texts from "\textit{Kavya Wajrayudhaya - Palamu Kotasa}" and the way \OldDataSetName{} and \DataSetName{} handled them.}
    \label{fig:multi-col-ex}
\end{figure}

During the post-processing of the OCR-extracted text in \OldDataSetName{}, it was discovered that the OCR engine has failed to separate columns in books that used column formatting, resulting in incorrect text rendering. This issue was resolved by using a two-column format to match the original book layout as mentioned in section~\ref{subsec:limits_sidiac}. However, there are challenges with this format when processing the documents, leading to the proposal of a single-column layout for \DataSetName{}, as discussed in subsection~\ref{subsec:other_post_steps}. This difference is highlighted in Figure~\ref{fig:multi-col-ex}, which provides an example illustrating the contrasting results of the two post-processing approaches used by the two datasets.

\begin{figure*}[!htbp]
    \centering
    \setlength{\fboxsep}{1pt} 
    \setlength{\fboxrule}{0.4pt} 
    \fbox{
        \includegraphics[width=0.98\textwidth]{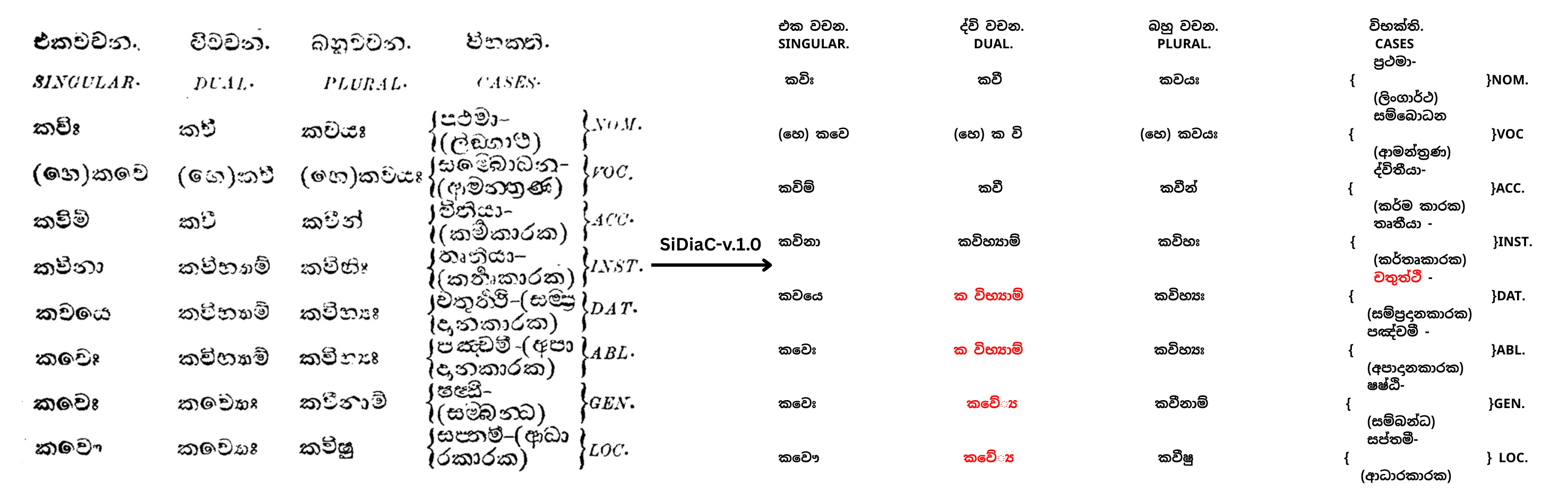}%
        }
    \caption{An example of content tables from "\textit{Sanskrutha Shabdhamalawa hewath Sanskrutha Nama Waranagilla}" in \OldDataSetName{}. Note that the words highlighted in red are malformed tokens.}
    \label{fig:content-tables}
\end{figure*}

The content tables in the documents of \OldDataSetName{} were designed to replicate their original format, resembling tables without actual table formatting, as discussed in section~\ref{subsec:limits_sidiac} and illustrated in Figure~\ref{fig:content-tables}. In \DataSetName{}, these tables were removed due to their diminished contextual relevance, which is further explained in subsection~\ref{subsec:other_post_steps}.

\begin{figure}[!htbp]
    \centering
    \setlength{\fboxsep}{1pt} 
    \setlength{\fboxrule}{0.4pt} 
    \fbox{
        \includegraphics[width=0.98\columnwidth]{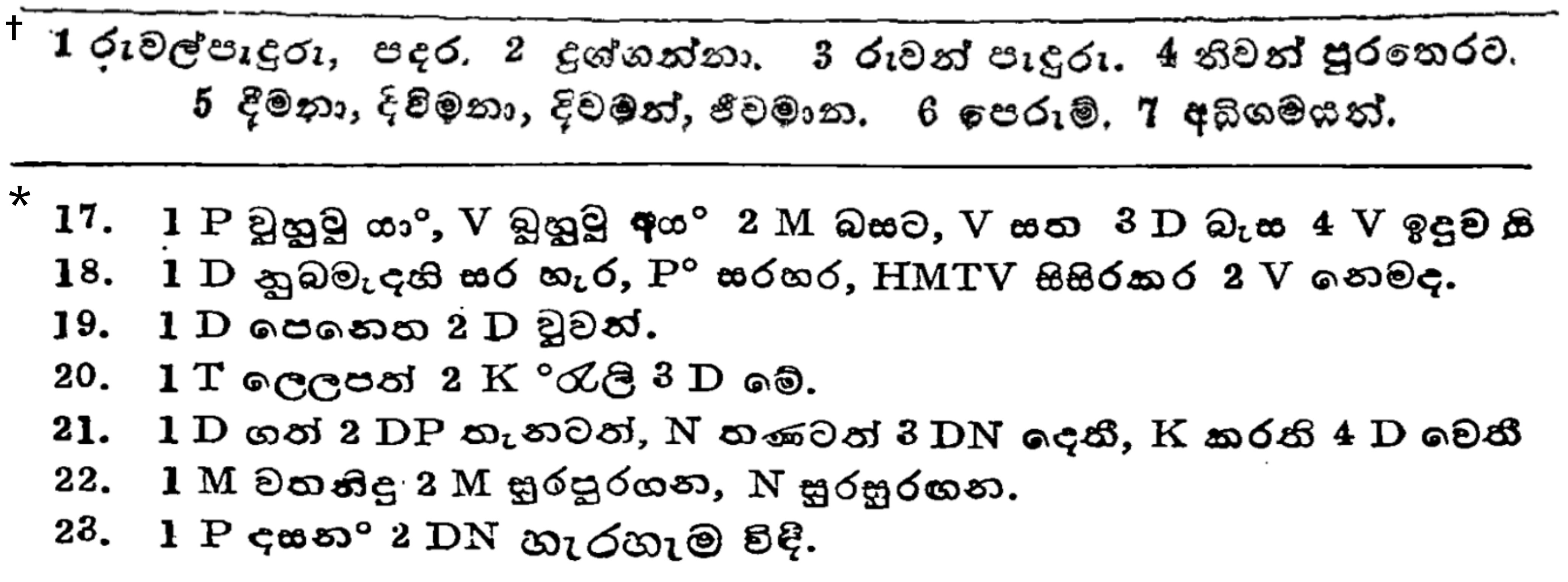}%
        }
    \caption{Examples of footnotes from "$^\dagger$\textit{Sadhdharma Rathnawaliya - Prathama Bagaya}" and "$^\star$\textit{Hansa Sandheshaya}".}
    \label{fig:footnotes}
\end{figure}

The footnotes in the books of \OldDataSetName{} were included in the text files without any indication that these specific phrases are not part of the main content. This inclusion disrupts the flow of the text, as discussed in section~\ref{subsec:limits_sidiac}. Two examples of such footnotes are illustrated in Figure~\ref{fig:footnotes}. To address this issue in \DataSetName{}, we have identified these text phrases (footnotes) and removed them to prevent any contextual disruption during text processing.

\section{Genre Analysis of \DataSetName}
\label{app:genre_analy}

The genre analysis for \DataSetName{} was conducted by dividing the published time frame of 1800 CE to 1955 CE into 20-year intervals, starting with the 1800-1820 bin and concluding with 1940-1955 (note that the dataset does not contain any data from 1955 to 1960 CE due to copyright restrictions as discussed in subsection~\ref{subsec:copyright_lang_filter}). This analysis, based on publication dates and genres, is summarised in Table~\ref{tab:dist_iss_sidiac}. The results revealed a right skew in the publication counts: only 13 out of \BookCount{} documents published between 1800 and 1880 were categorised: 5 in Religious, 4 in Poetry, 3 in Language, and 1 in History. The majority of books were released after 1880. The highest number of publications occurred in the period from 1880 to 1900, which included 63 books categorised under various secondary genres: 34 classified as Religious, 17 as Poetry, 3 as History, 2 as Language, 1 as Medical and 1 Unclassified. Overall, the analysis indicates that 141 of \BookCount{} books belong to either Poetry (54) or Religious (86) genres. This predominance may stem from the influences discussed in section~\ref{sec:eval}, particularly concerning the relationship between Buddhism and the impact of Sanskrit \textit{Kavya} traditions on Sinhala literature.



\begin{table*}[h!tb]
\centering
\resizebox{\textwidth}{!}{
\begin{tabular}{r|cc|ccccc|r}
\hline
& \multicolumn{2}{c|}{\textbf{Primary Category}} & \multicolumn{5}{c|}{\textbf{Secondary Category}} & \multirow{2}{*}{\textbf{Total}} \\ \cline{2-8}
 & \textbf{Fiction} & \textbf{Non-Fiction} & \textbf{History} & \textbf{Language} & \textbf{Medical} & \textbf{Poetry} & \textbf{Religious} & \\ \hline
\textbf{1800 - 1820} & 1 & 0 & 0 & 0 & 0 & 1 & 0 & 1 \\
\textbf{1820 - 1840} & 0 & 1 & 0 & 1 & 0 & 0 & 0 & 1 \\
\textbf{1840 - 1860} & 2 & 1 & 0 & 0 & 0 & 1 & 2 & 3 \\
\textbf{1860 - 1880} & 2 & 6 & 1 & 2 & 0 & 2 & 3 & 8 \\
\textbf{1880 - 1900} & 12 & 51 & 3 & 7 & 1 & 17 & 34 & *62 \\
\textbf{1900 - 1920} & 13 & 27 & 3 & 2 & 3 & 11 & 18 & *37 \\
\textbf{1920 - 1940} & 15 & 35 & 6 & 4 & 1 & 17 & 21 & *49 \\
\textbf{1940 - 1955} & 5 & 14 & 5 & 1 & 0 & 5 & 8 & 19 \\
\hline
\textbf{Total} & 50 & 135 & 18 & 17 & 5 & 54 & 86 & 185 \\
\hline
\end{tabular}}
\caption[]{\label{tab:dist_iss_sidiac}
Distribution of Books Across Issued Dates vs Genres in \DataSetName{}.\\
{\small *The total count for the secondary category between 1880 - 1900, 1900 - 1920, and 1920 - 1940 CE is 62, 37, and 50, respectively, while the overall number of books in those periods is 63, 40, and 50. This discrepancy arises because the books `\textit{Hithopadhesha Sannaya}', `\textit{Dhrawya Gunadharpana Sannaya}', `\textit{Asabandhi Sabandi}', `\textit{Ajuudha Neethiya}' and `\textit{Maadhanaa}' were not classified under any of the five secondary categories.}
}
\end{table*}

\begin{table*}
\centering
\begin{tabularx}{\textwidth}{r|ZZ|ZZZZZ|Z}
\hline
& \multicolumn{2}{c|}{\textbf{Primary Category}} 
& \multicolumn{5}{c|}{\textbf{Secondary Category}} &
\multirow{2}{*}{     \textbf{Total}}\\
\hhline{~-------~}
& \multicolumn{1}{c}{\textbf{Fiction}}& \multicolumn{1}{c|}{\textbf{Non-Fiction}} & \multicolumn{1}{c}{\textbf{History}} & \multicolumn{1}{c}{\textbf{Language}} & \multicolumn{1}{c}{\textbf{Medical}} & \multicolumn{1}{c}{\textbf{Poetry}} & \multicolumn{1}{c|}{\textbf{Religious}} & \\
\hline
\textbf{5th} & 0 & 1 & 0 & 0 & 1 & 0 & 0 & 1 \\
\textbf{12th} & 1 & 0 & 0 & 0 & 0 & 1 & 0 & 1 \\
\textbf{13th} & 1 & 12 & 1 & 3 & 0 & 1 & 8 & 13 \\
\textbf{14th} & 2 & 2 & 1 & 0 & 0 & 0 & 3 & 4 \\
\textbf{15th} & 4 & 4 & 0 & 2 & 0 & 3 & 3 & 8 \\
\textbf{16th} & 0 & 1 & 0 & 0 & 0 & 1 & 0 & 1 \\
\textbf{18th} & 0 & 4 & 0 & 1 & 1 & 0 & 2 & 4 \\
\textbf{19th} & 3 & 7 & 1 & 2 & 0 & 3 & 4 & 10 \\
\textbf{20th} & 5 & 12 & 2 & 2 & 0 & 6 & 6 & *16 \\
\hline
\textbf{Total} & 16 & 43 & 5 & 10 & 2 & 15 & 26 & 59 \\
\hline
\end{tabularx}
\caption[]{\label{tab:dist_cent_sidiac}
Distribution of Books Across Written Centuries vs Genres in \DataSetName{}-\texttt{filtered}.\\
{\small *The total count for the secondary category in the 20th century amounts to 16, while the overall number of books is 17. This discrepancy arises because the book `\textit{Hithopadhesha Sannaya}', which offers advice, was not classified under any of the five secondary categories.}
}
\end{table*}

We conducted genre analysis on \DataSetName{}-\texttt{filtered}, focusing on works written during different centuries. This analysis, which categorises documents by century-wise written dates and genre, is summarised in Table~\ref{tab:dist_cent_sidiac}. The 20th century contained the largest number of books, totalling 17, with 6 from both the Religious and Poetry genres, 2 each from History and Language and one unclassified. The 13th century followed with 13 books, of which 8 were Religious, and 3 were Language, while History and Poetry had one each. The 5th, 12th, and 16th centuries each had one book categorised as Medical (5th century) and as Poetry (12th and 16th centuries). The 19th century had 10 books, consisting of 4 Religious, 3 Poetry, 2 Language, and 1 History. All other centuries had fewer than 10 documents, with only the 15th century exceeding 5. Interestingly, the percentages of Fiction and Non-Fiction books in both \DataSetName{} and \DataSetName{}-\texttt{filtered} are similar, at 72.97\% and 72.88\% for Non-Fiction, respectively. 

\begin{table}[h!tb]
\centering
\resizebox{0.98\columnwidth}{!}{
\begin{tabular}{l|c|c|c|c|c|c|c}
\hline
 & \textbf{History} & \textbf{Language} & \textbf{Medical} & \textbf{Poetry} &  \textbf{Religious} & \textbf{Unclassified} & \textbf{Total} \\
\hline
\textbf{Fiction} & 2 & 0 & 0 & 41 & 5 & 2 & 50 \\
\textbf{Non-Fiction} & 16 & 17 & 5 & 13 & 81 & 3 & 135 \\ \hline
\textbf{Total} & 18 & 17 & 5 & 54 & 86 & 5 & 185 \\
\hline
\end{tabular}
}

\caption[]{\label{tab:freq_genre_1} Frequency distribution of books in \DataSetName{} categorised by primary and secondary genre classifications.}
\end{table}

\begin{table}[h!tb]
\centering
\resizebox{0.98\columnwidth}{!}{
\begin{tabular}{l|c|c|c|c|c|c|c}
\hline
 & \textbf{History} & \textbf{Language} & \textbf{Medical} & \textbf{Poetry} &  \textbf{Religious} & \textbf{Unclassified} & \textbf{Total} \\
\hline
\textbf{Fiction} & 0 & 0 & 0 & 12 & 4 & 0 & 16 \\
\textbf{Non-Fiction} & 5 & 10 & 2 & 3 & 22 & 1 & 43 \\ \hline
\textbf{Total} & 5 & 10 & 2 & 15 & 26 & 1 & 59 \\
\hline
\end{tabular}
}
\caption[]{\label{tab:freq_genre_2} Frequency distribution of books in \DataSetName{}-\texttt{filtered} categorised by primary and secondary genre classifications.}
\end{table}

\begin{table*}[!htb]
\centering
\resizebox{\textwidth}{!}{
\begin{tabular}{l|c|c|c|c|c|c}
\hline
\multirow{2}{*}{\textbf{Title}} & \multirow{2}{*}{\textbf{Issued Date}} & \multirow{2}{*}{\textbf{Author}} & \multirow{2}{*}{\textbf{Written Date}} & \multicolumn{2}{c|}{\textbf{Genre}} & \multirow{2}{*}{\textbf{OCR Confidence}$\uparrow$} \\
\cline{5-6}
&&&& \textbf{Primary} & \textbf{Secondary} &\\
\hline
Adhimasa Sangrahawa & 1903 & Madhampe Dhammathilaka Himi & 1850 - 1903 & Non-Fiction & Religious & 0.969200 \\\hline
Adhimasa Winishchaya & 1904 & Walikande Sri Sumangala Himi & 1850 - 1904 & Non-Fiction & Religious & 0.997100 \\\hline
Anagathawanshaya: Methe Budu Siritha & 1934 & \makecell{Watadhdhara Medhanandha Himi;\\ Siri Parakumabahu Wilgammula Sangaraja Himi} & 1325 - 1333 & Fiction & Religious & 0.999200 \\\hline
\makecell[l]{Ashoka Shilalipi saha \\Prathimakarana Winishchaya} & 1919 & D. E. Wickramasuriya & 1916 & Non-Fiction & History & 0.998900 \\\hline
Dhaham Sarana & 1931 & Unknown & 1220 - 1293 & Fiction & Religious & 0.989100 \\\hline
Dhaladha Pujawaliya & 1893 & Unknown & 1325 - 1333 & Non-Fiction & History & 0.997800 \\\hline
Dhampiya Atuwa Gatapadaya & 1932 & D.B. Jayathilaka & 1868 - 1932 & Non-Fiction & Religious & 0.924100 \\\hline
\makecell[l]{Dharma Pradheepikawa hewath \\Mahabodhiwansha Parikathawa} & 1906 & Unknown & 1187 - 1225 & Non-Fiction & Religious & 0.968200 \\\hline
Dhurwadhi Hardhaya Widharanaya & 1899 & Sri Dhanudhdharacharya & 1854 - 1899 & Non-Fiction & Religious & 0.991900 \\\hline
\makecell[l]{Gadaladeni Sannayai Prasidhdha wu \\Balawathare Purana Wyakyanaya} & 1877 & Hikkaduwe Sri Sumangala Himi & 1341 - 1408 & Non-Fiction & Language & 0.997000 \\\hline
Hansa Sandheshaya & 1953 & C.E. Godakumbure & 1457 - 1465 & Fiction & Poetry & 0.855300 \\\hline
Hithopadhesha Sannaya & 1884 & Waligama Sri Sumangala Himi & 1825 - 1905 & Non-Fiction & - & 0.987100 \\\hline
Jaanakeeharana & 1891 & Sri Kumaaradhaasa & 1201 - 1300 & Non-Fiction & History & 0.997600 \\\hline
Jubili Warnanawa & 1887 & John de Silva & 1857 - 1922 & Fiction & Poetry & 0.995700 \\\hline
Kathaluweeramanyawaadhaya & 1899 & Unknown & 1899 & Non-Fiction & Religious & 0.993600 \\\hline
Kavya Wajrayudhaya - Palamu Kotasa & 1889 & Engalthina Kumari & 1825 - 1893 & Fiction & Poetry & 0.925400 \\\hline
Kavyashekaraya & 1872 & Thotagamuwe Rahula Himi & 1408 - 1491 & Fiction & Poetry & 0.981300 \\\hline
Kudusika & 1894 & Unknown & 1270 - 1293 & Non-Fiction & Poetry & 0.998800 \\\hline
Kusajathaka Wiwaranaya (Prathama Bagaya) & 1932 & Munidhasa Kumarathunga & 1887 - 1932 & Non-Fiction & Religious & 0.970100 \\\hline
Lanka Maathaa & 1935 & S. Mahinda Himi & 1901 - 1951 & Fiction & Poetry & 0.856700 \\\hline
Liyanora Nadagama & 1936 & Unknown & 1852 - 1927 & Fiction & Poetry & 0.993500 \\\hline
Mage Malli & 1938 & G. H Perera & 1886 - 1938 & Fiction & Poetry & 0.865900 \\\hline
\makecell[l]{Maha Sanya sahitha Wishudhdhi\\ Maargaya - Chathurtha Baagaya} & 1955 & Unknown & 1266 - 1270 & Non-Fiction & Religious & 0.974500 \\\hline
\makecell[l]{Maha Sanya sahitha Wishudhdhi\\ Maargaya - Thruthiya Baagaya} & 1954 & Unknown & 1266 - 1270 & Non-Fiction & Religious & 0.968300 \\\hline
Moggallana Panchika Pradeepaya & 1896 & Unknown & 1070 - 1232 & Non-Fiction & Language & 0.991800 \\\hline
Muwadew da Wiwaranaya & 1949 & Munidhasa Kumarathunga & 1887 - 1944 & Non-Fiction & Religious & 0.871000 \\\hline
Nidhahase Manthraya & 1938 & S. Mahinda Himi & 1901 - 1938 & Non-Fiction & Poetry & 0.899700 \\\hline
Nikam Hakiyawa & 1941 & Munidhasa Kumarathunga & 1887 - 1941 & Fiction & Poetry & 0.893200 \\\hline
\makecell[l]{Nikaya Sangrahaya hewath\\ Shasanawatharaya} & 1922 & Unknown & 1390 & Non-Fiction & Religious & 0.975400 \\\hline
\makecell[l]{Okandapala Sannaya hewath \\Balawathara Liyana Sanna} & 1888 & Don Andhris Silva & 1760 - 1778 & Non-Fiction & Language & 0.988400 \\\hline
Paanadhuraa waadhaya & 1903 & P.A. Peris; P.J. Dhiyes & 1873 & Non-Fiction & Religious & 0.994500 \\\hline
Pansiya Panas Jathaka Potha & 1881 & Unknown & 1303 - 1333 & Fiction & Religious & 0.998700 \\\hline
Parani Gama & 1944 & Galpatha Kemanandha Himi & 1944 & Non-Fiction & History & 0.984600 \\\hline
Parawi Sandheshaya & 1873 & Unknown & 1430 - 1440 & Fiction & Poetry & 0.990200 \\\hline
\makecell[l]{Ruwanmal Nigantuwa hewath\\ Naamarathana Maalawa} & 1914 & \makecell{Hayaweni Sri Parakramabahu Rajathumaa;\\ D.P. de Alwis Wijesekara} & 1412 - 1467 & Non-Fiction & Language & 0.966200 \\\hline
Sadhdharma Rathnawaliya - Prathama Bagaya & 1930 & Dharmasena Himi & 1220 - 1293 & Non-Fiction & Religious & 0.996200 \\\hline
Sanna sahitha Abhisambodhi Alankaraya & 1897 & Waliwita Saranankara Sangaraja Himi & 1698 - 1778 & Non-Fiction & Religious & 0.998900 \\\hline
\makecell[l]{Sanna sahitha Kawsilumina hewath\\ Kusadhaawatha} & 1926 & \makecell{Kalikaala Sangeetha Sahithya;\\ Panditha Parakramabaahu Raja;\\ Madugalle Sidhdhartha Himi} & 1101 - 1200 & Fiction & Poetry & 0.942400 \\\hline
Sanna sahitha Salalihini Sandheshaya & 1859 & Unknown & 1450 & Fiction & Religious & 0.990900 \\\hline
\makecell[l]{Sanskrutha Shabdhamalawa hewath\\ Sanskrutha Nama Waranagilla} & 1876 & Rathmalane Dharmaloka Himi & 1828 - 1887 & Non-Fiction & Language & 0.967100 \\\hline
Sarartha Sangrahawa: Prathama Bhagaya & 1904 & Srimadh Budhdhadhasa Rajathuma & 398 - 426 & Non-Fiction & Medical & 0.999700 \\\hline
Sidath Sangaraawa - 1892 & 1892 & J P Amarasinghe & 1270 - 1293 & Non-Fiction & Language & 0.988300 \\\hline
Sidath Sangaraawa - 1954 & 1892 & Pathiraja Piruwan Himi; Ra. Thennakon & 1270 - 1293 & Non-Fiction & Language & 0.988300 \\\hline
Sinhala Upaasaka Janaalankaaraya & 1914 & Unknown & 1701 - 1800 & Non-Fiction & Religious & 0.991000 \\\hline
Sinhala Wyakaranaya enam Sidath Sangarawa & 1884 & Hikkaduwe Sri Sumangala Himi & 1827 - 1911 & Non-Fiction & Language & 0.988600 \\\hline
Sithiyam sahitha Mahiyangana Warnanawa & 1898 & Unknown & 1878 & Fiction & Poetry & 0.998900 \\\hline
Sithiyam sahitha Sadhdharmalankaraya & 1954 & Unknown & 1398 - 1410 & Non-Fiction & Religious & 0.981000 \\\hline
Sithiyam sahitha Sinhala Mahawanshaya & 1922 & D. H. S. Abayarathna & 1874 & Non-Fiction & History & 0.954900 \\\hline
Sithiyam sahitha Siyabas Maldhama & 1894 & Kirama Dhammanandha Himi & 1820 & Fiction & Poetry & 0.925600 \\\hline
Tibet Rate Budhudhahama & 1897 & Hendry S Olkat & 1832 - 1907 & Non-Fiction & Religious & 0.997300 \\\hline
Waidya Chinthamani Baishadhya Sangrahawa & 1909 & Unknown & 1706 - 1739 & Non-Fiction & Medical & 0.996500 \\\hline
Wibath Maldhama & 1906 & Kirama Dhammarama Himi & 1821 & Non-Fiction & Language & 0.998600 \\\hline
\makecell[l]{Wishudhdhi Maargaya - Prathama\\ Baagaya - 1949} & 1949 & Unknown & 1266 - 1270 & Non-Fiction & Religious & 0.953900 \\\hline
\makecell[l]{Wishudhdhi Maargaya Brahmawihaara\\ Nirdheshaathmaka Thruthiya Baagaya} & 1888 & \makecell{Budhdhagosha Himi;\\ Parakramabahu Rajathuma} & 1412 - 1467 & Non-Fiction & Religious & 0.991000 \\\hline
Wishudhdhi Margaya - Dhwitheeya Baagaya & 1851 & Unknown & 1266 - 1270 & Non-Fiction & Religious & 0.994400 \\\hline
\makecell[l]{Wishudhdhi Margaya - Prathama\\ Baagaya - 1897} & 1897 & Budhdhagosha Himi & 1266 - 1270 & Non-Fiction & Religious & 0.994700 \\\hline
\makecell[l]{Wistharaartha Granthipadha Wiwarnaya\\ sahitha Sinhala Wimaanawasthuuprakarana} & 1925 & \makecell{D.B.K. Gunathilaka;\\ Raajakarunaa Dissanayake} & 1469 - 1815 & Non-Fiction & Religious & 0.988200 \\\hline
\makecell[l]{Wyakarana Wiwarana hewath Sinhala\\ Bashawe Wyakaranaya} & 1937 & Munidhasa Kumarathunga & 1887 - 1937 & Non-Fiction & Language & 0.902900 \\\hline
Yoga Rathnaakaraya & 1930 & Unknown & 1501 - 1600 & Non-Fiction & Poetry & 0.946000 \\
\hline
\end{tabular}
}
\caption[]{\label{tab:meta_sidiac}
The metadata information for the \WrittenDateBookCount{} books used in the creation of \DataSetName{}-\texttt{filtered}.}
\end{table*}

As acknowledged in section~\ref{sec:eval}, there are five books that were not classified into any secondary genre. These are: `\textit{Hithopadhesha Sannaya}' (Issued: 1884, Written: 1825-1905), `\textit{Dhrawya Gunadharpana Sannaya}' (Issued: 1900-1920), `\textit{Asabandhi Sabandi}' (Issued: 1917), `\textit{Ajuudha Neethiya}' (Issued: 1924), and `\textit{Maadhanaa}' (Issued: 1917).
These books did not fit into any of the categories proposed by~\citet{jayatilleke2025sidiac} for \OldDataSetName{}, and we chose not to introduce new secondary genres only to accommodate just five out of \BookCount{} books.

It is important to note that primary and secondary genre classifications are independent. All \BookCount{} documents were classified as either `Non-fiction' or `Fiction'. Independently, all \BookCount{} documents, except for five books in \DataSetName{} and one book in \DataSetName{}-\texttt{filtered}, were classified in the genres of `History,' `Language,' `Medical,' `Poetry,' or `Religious,' as shown in Tables~\ref{tab:freq_genre_1} and \ref{tab:freq_genre_2}.


\section{Metadata Records of \DataSetName}
\label{app:eval_sdc}

The metadata for each document includes the title in Sinhala, the title in romanised form, the author's name in Sinhala (as indicated in the Digital Library of Natlib), the author's name in romanised form, genre categorised into primary (binary) and secondary (multi-class) genres, issued date, written date, and the OCR confidence, which is explained further in subsection \ref{subsec:metadata_creation}. A complete list of metadata information for \DataSetName{} can be found in the \texttt{GitHub}\footL[sidiacMetadata]{https://github.com/NeviduJ/SiDiaC-v.2.0/tree/main/Books_PDF\#readme} repository, while \DataSetName{}-\texttt{filtered} is presented in Table~\ref{tab:meta_sidiac}. 

\begin{table*}[h]
\centering
\resizebox{0.98\textwidth}{!}{
\begin{tabular}{l|c|c|c|c|c|c|c|c}
\hline
\textbf{Neighbour} & \textbf{Related Meaning(s)} & \textbf{13th} & \textbf{14th} & \textbf{15th} & \textbf{16th} & \textbf{18th} & \textbf{19th} & \textbf{20th} \\
\hline
\raisebox{-0.5ex}{%
    \includegraphics[height=1.25\fontcharht\font`\A]{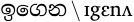}%
} & \multirow{7}{*}{Learn, Educate} &  &  &  & 2 &  &  &  \\ \cline{1-1}\cline{3-9}
\raisebox{-0.5ex}{%
    \includegraphics[height=1.45\fontcharht\font`\A]{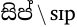}%
} &  &  &  &  &  &  &  & 2 \\\cline{1-1}\cline{3-9}
\raisebox{-0.5ex}{%
    \includegraphics[height=1.3\fontcharht\font`\A]{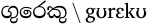}%
} &  &  &  &  &  &  &  & 3 \\\cline{1-1}\cline{3-9}
\raisebox{-0.5ex}{%
    \includegraphics[height=1.35\fontcharht\font`\A]{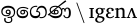}%
} &  &  &  &  &  &  & 1 &  \\\cline{1-1}\cline{3-9}
\raisebox{-0.5ex}{%
    \includegraphics[height=1.4\fontcharht\font`\A]{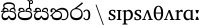}%
} &  &  &  &  &  &  &  & 1 \\\cline{1-1}\cline{3-9}
\raisebox{-0.5ex}{%
    \includegraphics[height=1.4\fontcharht\font`\A]{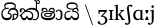}%
} &  & 1 &  &  &  &  &  &  \\\cline{1-1}\cline{3-9}
\raisebox{-0.5ex}{%
    \includegraphics[height=1.4\fontcharht\font`\A]{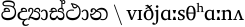}%
} &  & 1 &  &  &  &  &  &  \\ \hline
\raisebox{-0.5ex}{%
    \includegraphics[height=1.3\fontcharht\font`\A]{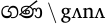}%
} & \multirow{6}{*}{Value, Math} & 5 &  &  &  &  & 1 &  \\\cline{1-1}\cline{3-9}
\raisebox{-0.5ex}{%
    \includegraphics[height=1.35\fontcharht\font`\A]{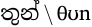}%
} &  & 1 &  &  &  &  &  & 1 \\\cline{1-1}\cline{3-9}
\raisebox{-0.5ex}{%
    \includegraphics[height=1.25\fontcharht\font`\A]{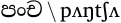}%
} &  &  &  &  &  & 3 &  &  \\\cline{1-1}\cline{3-9}
\raisebox{-0.5ex}{%
    \includegraphics[height=1.4\fontcharht\font`\A]{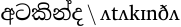}%
} &  &  &  &  &  &  & 1 &  \\\cline{1-1}\cline{3-9}
\raisebox{-0.5ex}{%
    \includegraphics[height=1.4\fontcharht\font`\A]{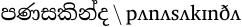}%
} &  &  &  &  &  &  & 1 &  \\\cline{1-1}\cline{3-9}
\raisebox{-0.5ex}{%
    \includegraphics[height=1.4\fontcharht\font`\A]{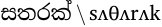}%
} &  &  &  &  & 1 &  &  &  \\\hline
\raisebox{-0.5ex}{%
    \includegraphics[height=1.25\fontcharht\font`\A]{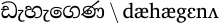}%
} & Thief &  &  &  &  &  & 1 &  \\\hline
\raisebox{-0.5ex}{%
    \includegraphics[height=1.2\fontcharht\font`\A]{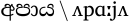}%
} & Hell & 1 &  & 1 &  &  &  &  \\\hline
\raisebox{-0.5ex}{%
    \includegraphics[height=1.2\fontcharht\font`\A]{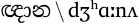}%
} & \multirow{6}{*}{Wisdom, Knowledge} &  &  &  &  & 4 &  &  \\ \cline{1-1}\cline{3-9}
\raisebox{-0.5ex}{%
    \includegraphics[height=1.3\fontcharht\font`\A]{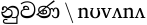}%
} &  &  &  &  &  & 1 &  &  \\\cline{1-1}\cline{3-9}
\raisebox{-0.5ex}{%
    \includegraphics[height=1.3\fontcharht\font`\A]{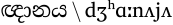}%
} &  &  &  & 1 &  &  &  &  \\\cline{1-1}\cline{3-9}
\raisebox{-0.5ex}{%
    \includegraphics[height=1.45\fontcharht\font`\A]{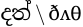}%
} &  &  &  &  &  &  & 1 &  \\\cline{1-1}\cline{3-9}
\raisebox{-0.5ex}{%
    \includegraphics[height=1.4\fontcharht\font`\A]{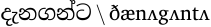}%
} &  &  &  &  & 1 &  &  &  \\\cline{1-1}\cline{3-9}
\raisebox{-0.5ex}{%
    \includegraphics[height=1.4\fontcharht\font`\A]{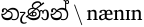}%
} & &  &  &  &  &  & 1 &  \\
 \hline
\end{tabular}

}
\caption[]{\label{tab:neigh_sath}
Frequency Distribution of Neighbour Words for "\raisebox{-0.5ex}{
\includegraphics[height=1.5\fontcharht\font`\A]{Figures/si_003.pdf}}" from the 13th to the 20th Century.}
\end{table*}

\begin{table*}[h]
\centering
\resizebox{0.98\textwidth}{!}{
\begin{tabular}{l|c|c|c|c|c|c|c|c}
\hline
\textbf{Neighbour} & \textbf{Related Meaning(s)} & \textbf{13th} & \textbf{14th} & \textbf{15th} & \textbf{16th} & \textbf{18th} & \textbf{19th} & \textbf{20th} \\
\hline
\raisebox{-0.5ex}{%
    \includegraphics[height=1.4\fontcharht\font`\A]{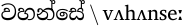}%
} & \multirow{8}{*}{\makecell[c]{Esteemed, Great, \\ Sacred}} & 1 & 6 &  &  &  & 2 & 2 \\ \cline{1-1}\cline{3-9}
\raisebox{-0.5ex}{%
    \includegraphics[height=1.4\fontcharht\font`\A]{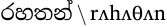}%
} &  &  & 2 &  &  &  & 1 & 4 \\ \cline{1-1}\cline{3-9}
\raisebox{-0.5ex}{%
    \includegraphics[height=1.4\fontcharht\font`\A]{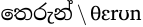}%
} &  & 1 & 3 &  &  &  & 1 &  \\ \cline{1-1}\cline{3-9}
\raisebox{-0.5ex}{%
    \includegraphics[height=1.22\fontcharht\font`\A]{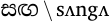}%
} &  &  & 2 & 1 &  &  &  & 2 \\ \cline{1-1}\cline{3-9}
\raisebox{-0.5ex}{%
    \includegraphics[height=1.4\fontcharht\font`\A]{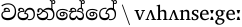}%
} &  & 1 & 3 &  &  &  &  &  \\ \cline{1-1}\cline{3-9}
\raisebox{-0.5ex}{%
    \includegraphics[height=1.4\fontcharht\font`\A]{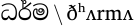}%
} &  & 1 & 1 &  &  &  &  & 1 \\ \cline{1-1}\cline{3-9}
\raisebox{-0.5ex}{%
    \includegraphics[height=1.3\fontcharht\font`\A]{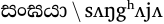}%
} &  &  &  & 2 &  & 1 &  &  \\ \cline{1-1}\cline{3-9}
\raisebox{-0.5ex}{%
    \includegraphics[height=1.4\fontcharht\font`\A]{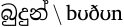}%
} &  & 2 &  &  &  &  &  & 1 \\ \hline
\raisebox{-0.5ex}{%
    \includegraphics[height=1.3\fontcharht\font`\A]{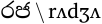}%
} & \multirow{11}{*}{Powerful, Strong} &  & 1 &  &  &  &  & 8 \\ \cline{1-1}\cline{3-9}
\raisebox{-0.5ex}{%
    \includegraphics[height=1.3\fontcharht\font`\A]{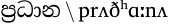}%
} & & 1 &  &  &  &  &  & 1 \\ \cline{1-1}\cline{3-9}
\raisebox{-0.5ex}{%
    \includegraphics[height=1.4\fontcharht\font`\A]{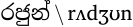}%
} &  &  &  &  &  &  &  & 4 \\ \cline{1-1}\cline{3-9}
\raisebox{-0.5ex}{%
    \includegraphics[height=1.4\fontcharht\font`\A]{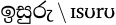}%
} & & 1 &  &  &  &  &  & 1 \\ \cline{1-1}\cline{3-9}
\raisebox{-0.5ex}{%
    \includegraphics[height=1.3\fontcharht\font`\A]{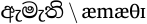}%
} &  &  &  &  &  &  &  & 2 \\ \cline{1-1}\cline{3-9}
\raisebox{-0.5ex}{%
    \includegraphics[height=1.3\fontcharht\font`\A]{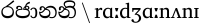}%
} &  &  &  &  &  &  &  & 2 \\ \cline{1-1}\cline{3-9}
\raisebox{-0.5ex}{%
    \includegraphics[height=1.3\fontcharht\font`\A]{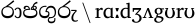}%
} & &  &  &  &  &  &  & 2 \\ \cline{1-1}\cline{3-9}
\raisebox{-0.5ex}{%
    \includegraphics[height=1.4\fontcharht\font`\A]{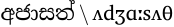}%
} &  &  & 2 &  &  &  &  &  \\ \cline{1-1}\cline{3-9}
\raisebox{-0.5ex}{%
    \includegraphics[height=1.3\fontcharht\font`\A]{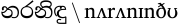}%
} &  &  &  &  &  &  &  & 2 \\ \cline{1-1}\cline{3-9}
\raisebox{-0.5ex}{%
    \includegraphics[height=1.35\fontcharht\font`\A]{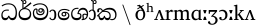}%
} &  &  &  &  &  &  &  & 2 \\ \cline{1-1}\cline{3-9}
\raisebox{-0.5ex}{%
    \includegraphics[height=1.3\fontcharht\font`\A]{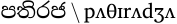}%
} & & 2 &  &  &  &  &  &  \\ \hline
\raisebox{-0.5ex}{%
    \includegraphics[height=1.35\fontcharht\font`\A]{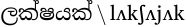}%
} & \multirow{5}{*}{\makecell[c]{Big, Large, \\Massive}} & 4 & 1 &  &  &  &  & 1 \\ \cline{1-1}\cline{3-9}
\raisebox{-0.5ex}{%
    \includegraphics[height=1.4\fontcharht\font`\A]{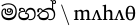}%
} &  &  & 1 & 1 &  &  &  &  \\ \cline{1-1}\cline{3-9}
\raisebox{-0.5ex}{%
    \includegraphics[height=1.45\fontcharht\font`\A]{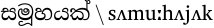}%
} &  &  &  & 1 &  &  &  & 1 \\ \cline{1-1}\cline{3-9}
\raisebox{-0.5ex}{%
    \includegraphics[height=1.3\fontcharht\font`\A]{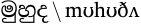}%
} &  & 2 &  &  &  &  &  &  \\ \cline{1-1}\cline{3-9}
\raisebox{-0.5ex}{%
    \includegraphics[height=1.25\fontcharht\font`\A]{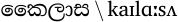}%
} &  &  &  & 2 &  &  &  &  \\
\hline
\end{tabular}
}
\caption[]{\label{tab:neigh_maha}
Frequency Distribution of Neighbour Words for "\raisebox{-0.5ex}{%
    \includegraphics[height=1.5\fontcharht\font`\A]{Figures/si_53.pdf}%
}" from the 13th to the 20th Century.}
\end{table*}

Based on the author counts, we conducted a frequency analysis on both \DataSetName{} and \DataSetName{}-\texttt{filtered}. During this analysis, we identified inconsistencies in the naming conventions of authors, which we standardised for accurate identification of unique counts. It is important to note that some books have multiple authors, with a maximum of three authors per book in \DataSetName{}. In \DataSetName{}, we found that out of \BookCount{} books, 69 had authors listed as \textit{`Unknown'}. The author with the highest number of books was "\textit{Munidhasa Kumarathunga}", who authored 4 books. Additionally, there were nine more authors who each had 2 books. The remaining 110 authors\footnote{\scriptsize There is an author named "\textit{D. H. S. Abayarathna}" who has written two books, and another author named "\textit{D. H. Stephen Abayarathna}". They are likely the same person, but we have kept the names as listed in the \texttt{Natlib} digital library metadata.} each contributed one book.
In \DataSetName{}-\texttt{filtered}, of the \WrittenDateBookCount{} books, 20 had authors listed as \textit{`Unknown'}. Once again, the author with the most books was "\textit{Munidhasa Kumarathunga}" with 4. There were also three authors with 2 books each, while 35 authors each had one book.

\section{Bag-of-Words Analysis}
\label{app:bow-analy}

Before conducting the diachronic comparison, a cross-century consistency filter was applied. Only target words that appeared in each of the seven centuries under examination (13th–20th century, excluding the 17th as no books were found in that century) were retained for analysis. This was achieved by computing the set intersection of vocabularies across all selected centuries. The resulting set of \ConsistentKeys{} consistent target words formed the basis for the cross-temporal collocate comparison, ensuring that any differences observed across centuries reflect genuine changes in language use rather than gaps caused by limited historical texts from certain periods.

The BoW analysis was conducted with a window span of ±10 words, as mentioned in section~\ref{sec:eval}. This analysis focused on two words known for their polysemies, based on the dictionary by~\citet{Soratha_Godage}. The two selected words, "\raisebox{-0.5ex}{%
\includegraphics[height=1.5\fontcharht\font`\A]{Figures/si_003.pdf}}" and "\raisebox{-0.5ex}{%
\includegraphics[height=1.5\fontcharht\font`\A]{Figures/si_53.pdf}}," were consistently present from the 13th to the 20th century in \DataSetName{}-\texttt{filtered}.

After extracting co-occurrence counts within a ±10 word window for each century, a longevity score was calculated for each collocate of words "\raisebox{-0.5ex}{%
\includegraphics[height=1.5\fontcharht\font`\A]{Figures/si_003.pdf}}" and "\raisebox{-0.5ex}{%
\includegraphics[height=1.5\fontcharht\font`\A]{Figures/si_53.pdf}}". This score is determined by multiplying the total co-occurrence count across the corpus by the number of centuries in which the collocate appeared at least once. This measure rewards collocates that are both high-frequency and temporally persistent, providing a data-driven criterion for identifying historically stable lexical associations with the target word.

Next, we sorted the collocates of the two words based on the descending order of longevity scores. We then selected the top 200 collocates and handpicked those that corresponded to our polysemous senses of "\raisebox{-0.5ex}{%
\includegraphics[height=1.45\fontcharht\font`\A]{Figures/si_003.pdf}}" and "\raisebox{-0.5ex}{%
\includegraphics[height=1.4\fontcharht\font`\A]{Figures/si_53.pdf}}" for qualitative study. The identified collocates are presented in Tables~\ref{tab:neigh_sath} and~\ref{tab:neigh_maha}.

The meanings of collocating words can be related to their associated categories, either directly or indirectly. For example, as illustrated in Table~\ref{tab:neigh_sath}, the word "\raisebox{-0.5ex}{%
    \includegraphics[height=1.45\fontcharht\font`\A]{Figures/si_003.pdf}%
}" is linked to the meanings "learn," "thief," "hell," and "wisdom," with all its neighbouring words being directly associated with these meanings. In contrast, the neighbour "\raisebox{-0.3ex}{%
    \includegraphics[height=1.4\fontcharht\font`\A]{Figures/si_081.pdf}%
}" is not directly related to the meanings of "value" and "math." However, in a mathematical context, "\raisebox{-0.3ex}{%
    \includegraphics[height=1.4\fontcharht\font`\A]{Figures/si_081.pdf}%
}" refers to "groups," which relate to quantities, thereby establishing an indirect connection to the meanings of "value" and "math".

When examining the word "\raisebox{-0.5ex}{%
    \includegraphics[height=1.4\fontcharht\font`\A]{Figures/si_53.pdf}%
}," as presented in Table~\ref{tab:neigh_maha}, it is evident that the neighbours categorised under the collective meanings of "esteemed," "great," and "sacred" are all directly connected to these terms. On the other hand, the neighbours "\raisebox{-0.4ex}{%
    \includegraphics[height=1.5\fontcharht\font`\A]{Figures/si_069.pdf}%
}\footL{https://en.wikipedia.org/wiki/Ajatashatru}" and "\raisebox{-0.3ex}{%
    \includegraphics[height=1.5\fontcharht\font`\A]{Figures/si_071.pdf}%
}\footL{https://en.wikipedia.org/wiki/Ashoka}," which are the names of well-known kings, are indirectly related to the meanings of "powerful" and "strong," as kings are typically associated with power. Additionally, the neighbour "\raisebox{-0.5ex}{%
    \includegraphics[height=1.3\fontcharht\font`\A]{Figures/si_077.pdf}%
}\footL{https://en.wikipedia.org/wiki/Mount_Kailash}," which refers to a mountain, is indirectly connected to the meaning "massive" because mountains in general (and specifically this particular mountain with a religious significance) are considered large.

\end{document}